%% file: neurips_2026.tex
\documentclass{article}

\usepackage[main,final]{neurips_2026}

\usepackage[utf8]{inputenc} 
\usepackage[T1]{fontenc}    
\usepackage{hyperref}       
\usepackage{url}            
\usepackage{booktabs}       
\usepackage{amsfonts}       
\usepackage{amsmath}
\usepackage{nicefrac}       
\usepackage{microtype}      
\usepackage{xcolor}         
\usepackage{graphicx}
\usepackage{color}
\usepackage{tabularray}
\usepackage{enumitem}
\usepackage{titletoc}

\title{Learning the Context of Errors: Black-Box Online Adaptation of Time Series Foundation Models}

\author{%
\textbf{Xilin Dai$^{1,2}$,\ \ Yiding Liu$^{1}$,\ \ Hongjie Xia$^{1}$,\ \ Yifan Hu$^{1}$,} \\
\textbf{Zewei Dong$^{1,*}$,\ \ Jiang-Ming Yang$^{1}$,\ \ Qiang Xu$^{2}$} \\
$^1$ Ant International \qquad $^2$ The Chinese University of Hong Kong \\
\texttt{\{xldai26, qxu\}@cse.cuhk.edu.hk} \\
\texttt{zewei.dong@ant-intl.com} \\
}

\begin{document}

\maketitle

\begin{abstract}
The rapid evolution of Time Series Foundation Models (TSFMs) has advanced zero-shot forecasting across diverse domains. Inspired by the current form of Large Language Models, future TSFMs may be offered as commercialized, closed-source application programming interface (API) services. However, many existing online adaptation methods still rely on white-box access for parameter fine-tuning or grey-box access for gradient backpropagation. This paradigm mismatch raises a question: \textit{In black-box online adaptation for TSFMs, what should we learn?} We answer this with an insight: the predictive errors of the base model are conditioned on both the input and output of the base model (i.e., the context of errors). To validate this insight, we propose \textbf{ORCA} (Online Residual Contextual Adaptation). We conduct extensive experiments across 5 state-of-the-art TSFMs and 8 datasets to demonstrate the effectiveness of our approach. Furthermore, through ablation studies, we quantitatively analyze the impact of different adapter learning hypotheses on the final adaptation performance in black-box online adaptation. Code available at https://github.com/Fifthky/ORCA.
\end{abstract}

\section{Introduction}
\label{sec:intro}

Time series forecasting is a cornerstone task across diverse domains, including energy management, traffic planning, and meteorology \citep{millerSurveyDeepLearning2024, liangFoundationModelsTime2024,dai_socgate_2025, huangTEFLPredictionResidualGuidedRolling2026}. Traditionally, time series forecasting has transitioned from classical statistical methods \citep{gardnerjr.ExponentialSmoothingState1985,piccoloDistanceMeasureClassifying1990} to deep learning architectures \citep{zhouInformerEfficientTransformer2021b,wuTimesNetTemporal2DVariation2022, zengAreTransformersEffective2023b}. Recently, inspired by the success of Large Language Models (LLMs), the forecasting paradigm is undergoing a shift towards Time Series Foundation Models (TSFMs) \citep{liangFoundationModelsTime2024, liu2026falconxtimeseriesfoundation}. Pre-trained on vast corpora of time series data spanning multiple domains, TSFMs demonstrate remarkable zero-shot capabilities when performing forecasting on unseen datasets\citep{aksuGIFTevalBenchmarkGeneral2024a, xu2026fideltshighfidelitymultimodalbenchmark, chen2026numericaltimeseriesunified}, including models such as the Chronos series \citep{ansariChronosLearningLanguage2024a,ansariChronos2UnivariateUniversal2025a} and the Moirai series \citep{wooUnifiedTrainingUniversal2024,liuMoirai20When2026,liuMoiraiMoEEmpoweringTime2025}.

\begin{figure}[t]
    \centering
    \includegraphics[width=\textwidth]{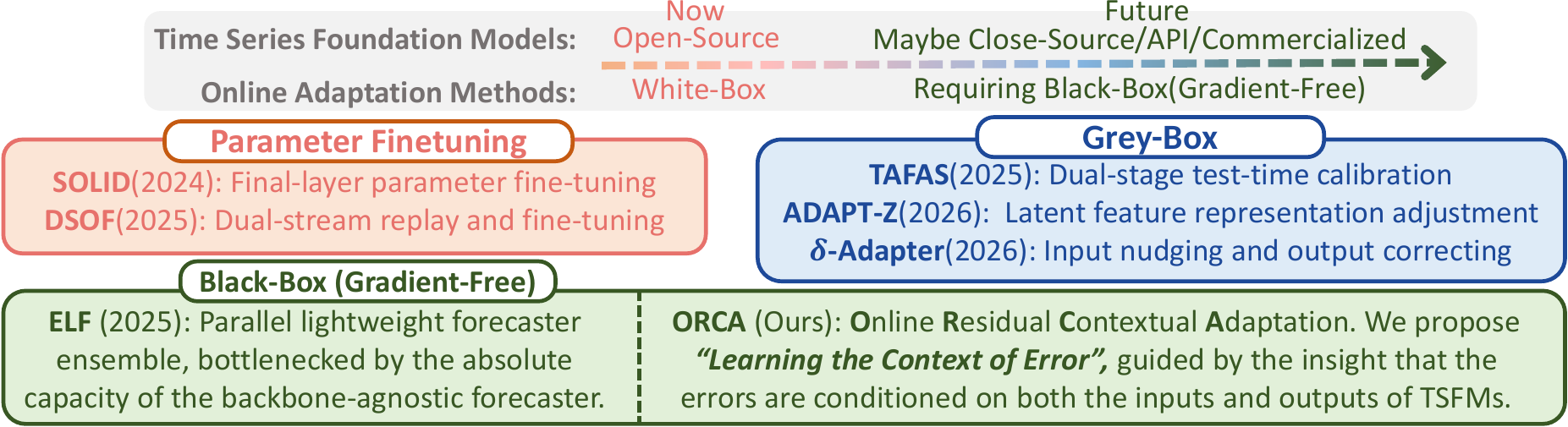}
    \caption{\textbf{Categorization of Time Series Online Adaptation Methods.} Depending on the access level to the base model, existing approaches are grouped into three paradigms. In the coming era of commercialized TSFM APIs, only the Black-Box paradigm could provide a feasible solution.}
    \label{fig:motivation}
\end{figure}

For TSFMs, online adaptation is particularly crucial: streaming data naturally suffers from temporal concept drift, and there remains a knowledge gap between the general pre-trained TSFMs and the specific dynamics of application scenarios \citep{zhangAddressingConceptShift2024, benechehabAdaPTSAdaptingUnivariate2025,dai2026positionuniversaltimeseries,leeAdaptingTimeSeries2026}. Depending on the required access level to the base model, existing online adaptation can be categorized into three paradigms (as illustrated in Figure \ref{fig:motivation}): (1) \textit{White-Box (Parameter Finetuning)}, exemplified by SOLID \citep{chenCalibrationTimeSeriesForecasting2024} and DSOF \citep{lauFastSlowStreams2024}, requires a local computation graph to explicitly modify internal weights or specific layers of the backbone network; (2) \textit{Grey-Box}, which includes TAFAS \citep{kimBattlingNonstationarityTime2025a}, $\delta$-Adapter \citep{liangForecastForecastPostProcessing2025}, and ADAPT-Z \citep{huangOnlineTimeSeries2025}, demands gradient backpropagation through the frozen foundation model to update input nudging or latent representations; (3) \textit{Black-Box}, such as ELF \citep{leeLightweightOnlineAdaption2025} and the Ada-Y variant of $\delta$-Adapter \cite{liangForecastForecastPostProcessing2025}, operates exclusively on the external interface.

Inspired by the success and commercial value of LLMs, future TSFMs may increasingly be offered as commercialized, closed-source API services \citep{xu2026internaldiagnosisexternalauditing}. Under this paradigm, users will only have black-box inference access. The strict API constraints render both White-Box and Grey-Box methods unfeasible. Consequently, the Black-Box paradigm emerges as the only viable solution. However, current explorations in this area are limited. ELF \citep{leeLightweightOnlineAdaption2025} performs a parallel lightweight forecaster ensemble. However, its performance is bottlenecked by the absolute capacity of this backbone-agnostic forecaster, and its contribution becomes marginal when the base model consistently dominates. Meanwhile, Ada-Y, an output-side variant of the $\delta$-Adapter \citep{liangForecastForecastPostProcessing2025}, solely relies on the base model's output, only capturing \textit{what} errors the model makes, without the input context necessary to understand \textit{when} those errors occur. 

This brings us to a question: \textit{\textbf{In black-box online adaptation for TSFMs, what should we learn?}} 
We argue that we should learn both what the errors are and, more crucially, when the base model makes these errors (i.e., \textbf{learning the context of errors}). Errors are not isolated noise; rather, they are conditioned on the inputs and outputs of the base model.  Specifically, we should model the conditional distribution of the errors given both the input sequences and the base model's predictions.

To actualize this insight, we propose \textbf{ORCA} (Online Residual Contextual Adaptation), a plug-and-play black-box adaptation framework for streaming TSFM API inference of time series. Given the input and the output of the base model, ORCA learns the context-aware error representation. To mitigate overfitting on noisy residuals, ORCA utilizes a \textit{Linear Adapter} with strong structural bias as its core component. To facilitate continuous learning, we design a buffer training mechanism with historical forgetting decay, alongside a predictive-space Bayesian loss. Furthermore, to utilize historical errors, we introduce a \textit{Boltzmann Router}. The router treats historical errors from both the base and the adapted predictions as Boltzmann energy states, dynamically deriving a confidence value to fuse them into a final combined output. The contributions of our work are summarized as follows:
\begin{itemize}[leftmargin=*]
    \item \textbf{Black-Box Online Adaptation for TSFMs:} We conduct black-box online adaptation analyses utilizing the latest generation of TSFMs, establishing a timely foundation for future research in commercialized API forecasting.

    \item \textbf{The ORCA Framework:} We introduce \textbf{ORCA}, a black-box online adaptation framework for TSFMs. ORCA utilizes a Linear Adapter with structural bias to provide context-aware residual corrections, integrating a buffer training mechanism with historical forgetting decay, a predictive-space Bayesian loss, and a dynamic Boltzmann Router.
    
    \item \textbf{Contextual Error Modeling and ``What to Learn'' :} We propose that adapters should learn the context of errors, both \textit{what} the errors are and \textit{when} the base model makes them. Through ablation studies on the adapter's input configurations, we quantitatively analyze the impact of different learning hypotheses.

\end{itemize}

\section{Related works}
\label{sec:related}
\subsection{Deep learning for time series forecasting}
Historically, the evolution of time series forecasting has transitioned from classical statistical methods \citep{gardnerjr.ExponentialSmoothingState1985,piccoloDistanceMeasureClassifying1990} to deep learning architectures, including Recurrent Neural Networks (RNNs), Convolutional Neural Networks (CNNs) \citep{connorRecurrentNeuralNetworks1994,hochreiterLongShortTermMemory1997b,laiModelingLongShortTerm2018,dai_socnet_2025}, and subsequently, advanced structures like Transformers \citep{zhouInformerEfficientTransformer2021b,wuTimesNetTemporal2DVariation2022, liuNonstationaryTransformersExploring2022, nieTimeSeriesWorth2022a, liuITransformerInvertedTransformers2023, wangTimeXerEmpoweringTransformers2024} and state-space models \citep{ahamedTimeMachineTimeSeries2024,maMambaFoundationModel2024}. Researchers have also questioned the necessity of complex attention mechanisms, showing that simple linear models \citep{zengAreTransformersEffective2023b, xuFITSModelingTime2023b, daiSamplesScenariosNew2025a} can achieve competitive results.

\subsection{Time series foundation models (TSFMs)}

Driven by the success of pre-training in natural language and vision, the community has begun to focus on Time Series Foundation Models (TSFMs) \citep{liangFoundationModelsTime2024,meyerTimeSeriesFoundation2025,millerSurveyDeepLearning2024}, which aim to provide universal, zero-shot forecasting capabilities. Popular paradigms involve reprogramming existing LLMs for time series or pre-training transformers from scratch with cross-domain time series data, such as Time-LLM \citep{jinTimeLLMTimeSeries2023}, Chronos family \citep{ansariChronosLearningLanguage2024a,ansariChronos2UnivariateUniversal2025a}, Lag-Llama \citep{rasulLagLlamaFoundationModels2024a}, and other architectures \citep{dasDecoderonlyFoundationModel2024,zhouOneFitsAll2023a, chenVisionTSVisualMasked2025}. Concurrent developments have also introduced TimeGPT-1 \cite{garzaTimeGPT12024}, the Moirai series (Moirai 1.0, 2.0, and Moirai-MoE) \cite{wooUnifiedTrainingUniversal2024,liuMoirai20When2026,liuMoiraiMoEEmpoweringTime2025}, the Timer family (Timer, Timer-S1) \cite{liuTimerGenerativePretrained2024,liuTimerS1BillionScaleTime2026}, Sundial \cite{liuSundialFamilyHighly2025a}, TiRex \cite{auerTiRexZeroShotForecasting2025}, and lightweight models like Tiny Time Mixers (TTM) \cite{ekambaramTinyTimeMixers2024}. The static nature of these models leaves them vulnerable to temporal concept drifts and the knowledge gap between pre-training data and real-world applications.

\subsection{Online learning in time series}
Real-world time series data is intrinsically non-stationary, frequently experiencing concept drift \citep{besnardContinualLearningTime2024,zhangAddressingConceptShift2024}. To mitigate catastrophic forgetting \citep{kirkpatrickOvercomingCatastrophicForgetting2017} and maintain plasticity \citep{dohareMaintainingPlasticityDeep2024,aoContinualDeepLearning2023,verwimpContinualLearningApplications2023,behrouzNestedLearningIllusion2025b}, continual and online learning methods update models sequentially as new data arrives. Classical deep online forecasting methods include FSNet \citep{phamLearningFastSlow2022} and OneNet \citep{zhangOneNetEnhancingTime2023}, as independent online evolving forecasters. When discussing time series online adapters, depending on the required access level to the base model, existing approaches can be categorized into three paradigms: (1) \textit{White-Box (Parameter Finetuning)}, such as SOLID \citep{chenCalibrationTimeSeriesForecasting2024} and DSOF \citep{lauFastSlowStreams2024}; (2) \textit{Grey-Box}, which includes TAFAS \citep{kimBattlingNonstationarityTime2025a}, $\delta$-Adapter \citep{liangForecastForecastPostProcessing2025}, ADAPT-Z \citep{huangOnlineTimeSeries2025}, PETSA \citep{medeirosAccurateParameterEfficientTestTime2025a} and DynaTTA \citep{groverShiftAwareTestTime2025}; (3) \textit{Black-Box}, such as ELF \citep{leeLightweightOnlineAdaption2025} and the Ada-Y variant of $\delta$-Adapter \citep{liangForecastForecastPostProcessing2025}. Additionally, broader explorations in other domains like traffic and spatial-temporal forecasting have introduced methods like FORESEE \citep{huangLearningYesterdaysError2026}, and ADCSD \citep{guoOnlineTestTimeAdaptation2025}. However, current literature specifically exploring online adaptation for TSFMs mainly includes TAFAS \cite{kimBattlingNonstationarityTime2025a}, ELF \cite{leeLightweightOnlineAdaption2025}, $\delta$-Adapter \cite{liangForecastForecastPostProcessing2025}, and remain insufficiently explored.

\subsection{Comparison with online learning in natural language processing}
While fine-tuning and online learning for LLMs such as \cite{huLoRALowRankAdaptation2021,bidermanLoRALearnsLess2024,houlsbyParameterEfficientTransferLearning2019,pfeifferMADXAdapterBasedFramework2020, liPrefixTuningOptimizingContinuous2021, xu2026contextualagenticmemorymemo, hu2026tomorrowtodayselfevolvingpolicies} have been widely analyzed, the online adaptation of TSFMs presents a different paradigm. The primary divergence lies in the availability of feedback during online inference. In typical natural language processing (NLP) applications, exact ground truth is rarely immediately accessible during test-time generation. In contrast, in time series forecasting, the ground truth reveals itself as time progresses, and this distinction motivates our \textbf{ORCA} methodology.

\begin{figure}[t]
    \centering
    \includegraphics[width=\textwidth]{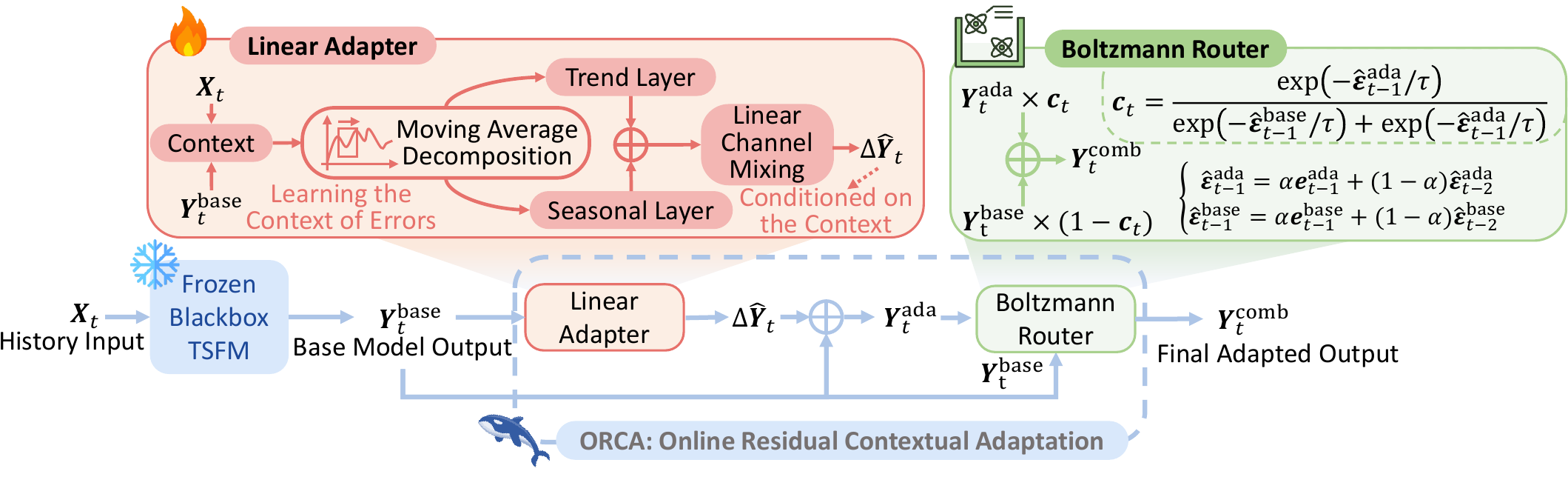}
    \caption{The overall architecture of ORCA. The history input $\boldsymbol{X}_t$ is fed into a frozen black-box TSFM to obtain the base output $\boldsymbol{Y}_t^{\mathrm{base}}$. The Linear Adapter learns to generate the residual via Moving Average Decomposition, followed by Trend and Seasonal Layers. The Boltzmann Router dynamically calculates the confidence score $\boldsymbol{c}_t$ to mix $\boldsymbol{Y}_t^{\mathrm{ada}}$ and $\boldsymbol{Y}_t^{\mathrm{base}}$ into the final adapted output $\boldsymbol{Y}_t^{\mathrm{comb}}$.}
    \label{fig:structure}
\end{figure}

\section{Methodology}
\label{sec:methodology}
\subsection{Problem formulation}
Consider a streaming time series forecasting scenario where data arrives sequentially. At each time step $t$, we observe a historical input matrix $\mathbf{X}_t \in \mathbb{R}^{D \times L}$, where $D$ is the number of channels (variates) and $L$ is the look-back window length. The objective is to forecast the future values over a horizon $H$. A pre-trained, frozen black-box TSFM takes $\mathbf{X}_t$ as input and generates a base prediction $\mathbf{Y}_t^{\mathrm{base}} \in \mathbb{R}^{D \times H}$. Due to the non-stationarity of real-world environments and the knowledge gap between the pre-trained TSFM and the application scenario, the base model's prediction deviates from the ground truth $\mathbf{Y}_t^{\mathrm{GT}} \in \mathbb{R}^{D \times H}$. We define the true prediction error as $\mathbf{E}_{\mathrm{obs}, t} = \mathbf{Y}_t^{\mathrm{GT}} - \mathbf{Y}_t^{\mathrm{base}}$. Because the base TSFM is treated as a black box (e.g., accessed via an API without access to internal gradients), we can only apply output-side corrections. Our goal is to learn an adapter function $f_{\theta}$ with parameters $\theta$ that predicts the residual $\Delta\hat{\mathbf{Y}}_t$ to refine the base prediction, ultimately yielding an adapted output $\mathbf{Y}_t^{\mathrm{ada}} = \mathbf{Y}_t^{\mathrm{base}} + \Delta\hat{\mathbf{Y}}_t$.

\subsection{Context conditioned learning}
ORCA aims to precisely capture the context in which base model errors occur. Conceptually, we aim to model the conditional error distribution $P(\mathbf{E}_t \mid \mathbf{X}_t, \mathbf{Y}_t^{\mathrm{base}})$, where $\mathbf{E}_t$ denotes the true residual matrix at time $t$. Instead of relying on explicit probabilistic generative modeling, we pull this concept back into a deterministic forecasting paradigm. Let the contextual condition be denoted as $\mathbf{C}_t = [\mathbf{X}_t, \mathbf{Y}_t^{\mathrm{base}}]$, and our Linear Adapter as a deterministic mapping $\Delta\hat{\mathbf{Y}}_t = f(\mathbf{C}_t)$. By optimizing the Mean Squared Error (MSE) loss during adaptation, we are minimizing the expected prediction risk. As derived in Appendix \ref{app:context_conditioned}, minimizing this risk mathematically dictates that the optimal deterministic mapping $f^*(\mathbf{C}_t)$ is exactly the conditional expectation:
\begin{equation}
    f^*(\mathbf{C}_t) = \mathbb{E} [\mathbf{E}_t \mid \mathbf{X}_t, \mathbf{Y}_t^{\mathrm{base}}]
\end{equation}
This standard property of MSE optimization formalizes our objective: it allows us to rigorously translate the probabilistic modeling of $P(\mathbf{E}_t \mid \mathbf{X}_t, \mathbf{Y}_t^{\mathrm{base}})$ into a purely deterministic residual regression task. Guided by this, our Linear Adapter takes the concatenation of the historical input $\mathbf{X}_t$ and the base prediction $\mathbf{Y}_t^{\mathrm{base}}$ as its context, ensuring the model conditions on the relevant variables to estimate this expected residual.

\subsection{Linear adapter}

As illustrated in Figure \ref{fig:structure}, the Linear Adapter utilizes a linear architecture. Because the residual signal is highly noisy, we deliberately choose a linear architecture with strong inductive bias to provide the regularization and avoid overfitting. Specifically, the contextual input $\mathbf{C}_t$ is first processed through a Moving Average Decomposition block, which separates the temporal dynamics into a trend component and a seasonal component. These components are independently processed by a Trend Layer and a Seasonal Layer. Finally, Linear Channel Mixing layers across the channel dimension capture cross-variate dependencies, outputting the predicted residual. The adapter output is then added to the base prediction to form the adapted output: $ \mathbf{Y}_t^{\mathrm{ada}} = \mathbf{Y}_t^{\mathrm{base}} + \Delta\hat{\mathbf{Y}}_t$.

\subsection{Boltzmann routing mechanism}
\label{sec:boltzmann}
To reduce the risk of online adaptation underperforming the base model, we introduce the \textbf{Boltzmann Router}. Drawing inspiration from statistical mechanics, where the Boltzmann distribution defines the probability of a system being in a specific state as a function of its energy, we analogously treat the smoothed predictive error as the energy of a model state. A lower error corresponds to a lower energy state, thus yielding a higher routing probability (confidence score). Let $\mathbf{e}_t^{\mathrm{base}}$ and $\mathbf{e}_t^{\mathrm{ada}}$ denote the instantaneous absolute error vectors (across $D$ channels) for the base model and the adapter, respectively. We track their exponential moving averages (EMA) to estimate the smoothed errors (energies):
\begin{equation}
    \begin{cases}
        \mathbf{\hat{\varepsilon}}_{t-1}^{\mathrm{ada}} = \alpha \mathbf{e}_{t-1}^{\mathrm{ada}} + (1 - \alpha) \mathbf{\hat{\varepsilon}}_{t-2}^{\mathrm{ada}} \\
        \mathbf{\hat{\varepsilon}}_{t-1}^{\mathrm{base}} = \alpha \mathbf{e}_{t-1}^{\mathrm{base}} + (1 - \alpha) \mathbf{\hat{\varepsilon}}_{t-2}^{\mathrm{base}}
    \end{cases}
\end{equation}
where $\alpha \in (0,1)$ is the momentum coefficient. The channel-wise routing confidence vector $\mathbf{c}_t \in \mathbb{R}^D$ is then calculated via a Boltzmann softmax function with temperature $\tau$:
\begin{equation}
    \mathbf{c}_t = \frac{\exp(-\mathbf{\hat{\varepsilon}}_{t-1}^{\mathrm{ada}} / \tau)}{\exp(-\mathbf{\hat{\varepsilon}}_{t-1}^{\mathrm{base}} / \tau) + \exp(-\mathbf{\hat{\varepsilon}}_{t-1}^{\mathrm{ada}} / \tau)}
\end{equation}

During inference, this confidence vector $\mathbf{c}_t$ gates the integration of the adapter's correction:
\begin{equation}
    \mathbf{Y}_t^{\mathrm{comb}} = \mathbf{Y}_t^{\mathrm{base}} \odot (\mathbf{1} - \mathbf{c}_t) + \mathbf{Y}_t^{\mathrm{ada}} \odot \mathbf{c}_t
\end{equation}
where $\odot$ denotes element-wise multiplication broadcasted along the sequence dimension. From the perspective of online learning, this weighting mechanism connects to Prediction with Expert Advice and the Multiplicative Weights Update (MWU) framework \citep{freundDecisionTheoreticGeneralizationOnLine1997}, acting to track the best expert in non-stationary environments \citep{herbsterTrackingBestExpert1998}. During online training, the channel-mean of this confidence vector, $\bar{c}_t = \mathrm{mean}(\mathbf{c}_t)$, naturally acts as the prior regularization weight in our subsequent Bayesian loss function. A detailed theoretical analysis, including the regret bound of this Boltzmann routing mechanism, is provided in the Appendix \ref{app:boltzmann}.

\subsection{Predictive-space Bayesian update}
\label{sec:bayesian}

A pivotal challenge in online learning is the plasticity-stability dilemma. Instead of imposing complex regularization on model parameters $\theta$, we formulate the optimization as a Bayesian update directly in the predictive space. We define the belief over the conditional error distribution $P(\mathbf{E}_t \mid \mathbf{C}_t)$. At step $t$, we assume the observation likelihood follows an isotropic Gaussian distribution centered at the true error with noise variance $\sigma_{\mathrm{obs}}^2$, denoted as $P(\mathbf{E}_{\mathrm{obs}, t} \mid \mathbf{E}_t, \mathbf{C}_t) \sim \mathcal{N}(\mathbf{E}_t, \sigma_{\mathrm{obs}}^2 \mathbf{I})$, where the observed error is $\mathbf{E}_{\mathrm{obs}, t} = \mathbf{Y}_t^{\mathrm{GT}} - \mathbf{Y}_t^{\mathrm{base}}$. To prevent catastrophic forgetting, we construct a prior distribution based on the prediction of the previous cycle's model, $\theta_{\mathrm{prior}}$. The prior is defined as $P(\mathbf{E}_t \mid \mathbf{C}_t, \mathcal{H}_{t-1}) \sim \mathcal{N}(\mathbf{E}_{\mathrm{prior}, t}, \sigma_{\mathrm{prior}}^2 \mathbf{I})$, where $\mathcal{H}_{t-1}$ denotes historical information, $\mathbf{E}_{\mathrm{prior}, t} = f_{\theta_{\mathrm{prior}}}(\mathbf{C}_t)$ denotes the expected error predicted by the historical model, and $\sigma_{\mathrm{prior}}^2$ represents the variance of this prior belief.

By applying Bayes' theorem and performing maximum a posteriori (MAP) estimation, minimizing the negative log-posterior is mathematically equivalent to minimizing the $L_2$ distance to both the observation and the prior, weighted by the precision ratio $\lambda_t = \sigma_{\mathrm{obs}}^2 / \sigma_{\mathrm{prior}}^2$. As expanded in Appendix \ref{app:bayesian}, by transforming the errors back into the target time series space, we arrive at the Bayesian-inspired loss function defined on a training snapshot $s_t = \{\mathbf{X}_t, \mathbf{Y}_t^{\mathrm{base}}, \mathbf{Y}_t^{\mathrm{GT}}\}$:
\begin{equation}
    \mathcal{L}(s_t, \theta) = \frac{1}{H \times D} \left( \|\mathbf{Y}_t^{\mathrm{GT}} - \mathbf{Y}_t^{\mathrm{ada}}\|_F^2 + \bar{c}_t \|\mathbf{Y}_t^{\mathrm{ada}} - \mathbf{Y}_t^{\mathrm{prior}}\|_F^2 \right)
\end{equation}
where $\|\cdot\|_F$ denotes the Frobenius norm, $\mathbf{Y}_t^{\mathrm{prior}} = \mathbf{Y}_t^{\mathrm{base}} + \mathbf{E}_{\mathrm{prior}, t}$, and the Linear Adapter output constructs $\mathbf{Y}_t^{\mathrm{ada}} = \mathbf{Y}_t^{\mathrm{base}} + \Delta\hat{\mathbf{Y}}_t$. The prior prediction is a stop-gradient target generated by the frozen adapter snapshot from the previous training cycle; gradients update only the current adapter. 

We substitute the precision ratio $\lambda_t$ with $\bar{c}_t = \mathrm{mean}(\mathbf{c}_t)$, the scalar channel-mean derived from the Boltzmann Router. Solving for the optimal adapter output $\Delta\hat{\mathbf{Y}}_t^*$ gives:
\begin{equation}
    \Delta\hat{\mathbf{Y}}_t^* = \frac{1}{1+\bar{c}_t} \mathbf{E}_{\mathrm{obs}, t} + \frac{\bar{c}_t}{1+\bar{c}_t} \mathbf{E}_{\mathrm{prior}, t}
    \label{eq:bayesian_optimal}
\end{equation}
This formulation corresponds exactly to the analytical mean of the product of two Gaussian distributions, as derived in Appendix \ref{app:bayesian}. As prior variance is computationally impractical and boundless, $\bar{c}_t$ serves as a stable surrogate. Practically, the exponential moving average of historical errors serves as a first-order proxy for localized predictive variance, and the Boltzmann Softmax smoothly maps this into a relative precision score. When the adapter performs well ($\bar{c}_t \to 1$), the prior precision is high, anchoring the model to historical beliefs. Conversely, the increase of adapter errors causes $\bar{c}_t \to 0$ to drop the prior, forcing the adapter to quickly digest new patterns.

\subsection{Inference and online training pipeline}
\label{sec:pipeline}
\begin{figure}[h]
    \centering
    \includegraphics[width=\textwidth]{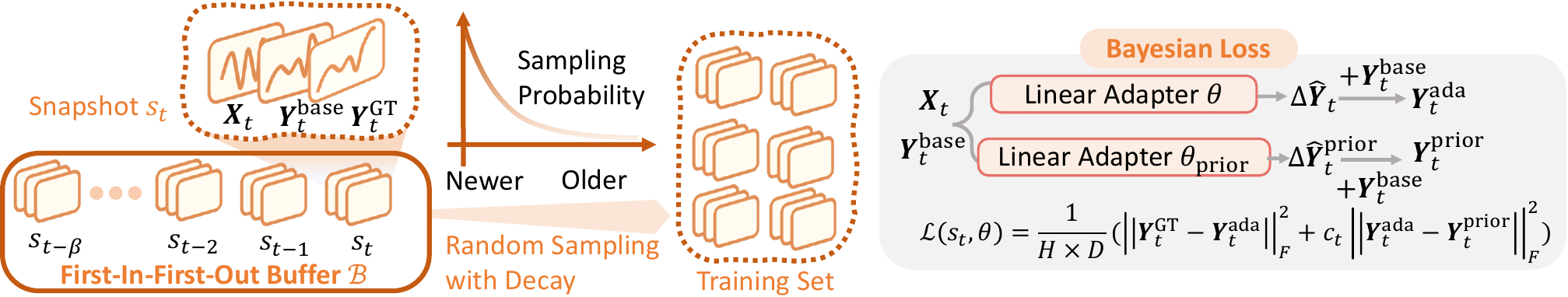}
    \caption{The online training pipeline of ORCA. Snapshots $s_t$ containing $\{\boldsymbol{X}_t, \boldsymbol{Y}_t^{\mathrm{base}}, \boldsymbol{Y}_t^{\mathrm{GT}}\}$ are stored in a FIFO Buffer. Random sampling with an exponential decay probability prioritizes recent samples. The Bayesian Loss simultaneously aligns $\boldsymbol{Y}_t^{\mathrm{ada}}$ with the ground truth $\boldsymbol{Y}_t^{\mathrm{GT}}$ and anchors it to the prior prediction $\boldsymbol{Y}_t^{\mathrm{prior}}$ generated by the frozen, stop-gradient model copy $\theta_{\text{prior}}$.}
    \label{fig:train}
\end{figure}

ORCA operates through a step-by-step inference and periodic cycle-training pipeline, utilizing a First-In-First-Out (FIFO) replay buffer $\mathcal{B}$ (Figure \ref{fig:train}). During continuous inference at each time step $t$, the Linear Adapter conditions on the observable context $\mathbf{C}_t$ to generate the residual correction $\Delta\hat{\mathbf{Y}}_t$. Once the true horizon becomes fully observable, we construct a training snapshot $s_t$ and push it into the FIFO buffer $\mathcal{B}$. ORCA executes cycle training periodically every horizon $H$ steps. During each training cycle, we draw batches from the buffer using random sampling with an exponential decay probability. This mechanism assigns a higher selection likelihood to more recent snapshots. Consequently, it guarantees that the Linear Adapter maintains high plasticity towards recent distribution shifts, while retaining a sufficient proportion of historical anchor points to stabilize the Bayesian prior. Once a training cycle is complete, the current adapter parameters $\theta$ are frozen to update the delayed prior model $\theta_{\mathrm{prior}}$, preparing the Bayesian loss anchor for the subsequent cycle.

\section{Experiments}
\label{sec:experiments}

\subsection{Experimental setup}
\label{subsec:experimental_setup}

\textbf{Datasets.} We evaluate our proposed ORCA framework on eight widely used real-world time series benchmarks: ETTh1, ETTh2, ETTm1, ETTm2, Electricity \citep{zhouInformerEfficientTransformer2021b}, Exchange \citep{laiModelingLongShortTerm2018}, Weather, and Traffic \citep{wuAutoformerDecompositionTransformers2021b}. The full set of eight datasets is employed for the main experiments to compare ORCA against various baselines, as in previous work \citep{liangForecastForecastPostProcessing2025}. However, given the computational cost associated with online evaluation across all datasets for all base models, our ablation studies and hyperparameter sensitivity analyses are conducted on a subset of six datasets. This subset excludes the high-dimensional Traffic and Electricity datasets, which aligns with the evaluation protocols adopted by prior works \citep{kimBattlingNonstationarityTime2025a}.

\textbf{Base Models.} Existing online adaptation works have primarily conducted experiments on first-generation time series foundation models \cite{kimBattlingNonstationarityTime2025a, leeLightweightOnlineAdaption2025, liangForecastForecastPostProcessing2025}. Consequently, there remains a lack of evaluations on the latest base models that currently dominate public leaderboards such as GIFT-Eval \citep{aksuGIFTevalBenchmarkGeneral2024a} and fev-bench \citep{shchurFevbenchRealisticBenchmark2026}. To bridge this gap, we select five recent (released between 2025 and 2026), popular, and high-performing base models for our evaluation: Chronos-2 \cite{ansariChronos2UnivariateUniversal2025a}, Moirai 2.0 \cite{liuMoirai20When2026}, TiRex \cite{auerTiRexZeroShotForecasting2025}, TimesFM-2.5 (the version released in September 2025) \cite{dasDecoderonlyFoundationModel2024}, and Sundial \cite{liuSundialFamilyHighly2025a}. Across all selected models, the input look-back window length is uniformly set to $L = 520$, and the testing horizons are set to $H \in \{30, 96, 336\}$ \citep{leeLightweightOnlineAdaption2025}. This setting follows ELF and provides a common input/output interface across heterogeneous frozen TSFMs, covering short-, medium-, and long-range prediction. Unlike full-shot forecasting, where a separate model is trained on each dataset, this protocol controls the comparison across frozen backbones. Unless specified otherwise, results in this paper represent the average performance across these three horizons, a common practice in prior work \citep{liangForecastForecastPostProcessing2025}. While TSFMs output probabilistic forecasts, our framework focuses on operating on the median of the quantile or sampled outputs.

\textbf{Other Settings.} The primary objective of this paper is to investigate the potential for performance enhancement when the foundation model operates strictly as a black box (e.g., accessed via a cloud API). However, because the most recognized and highly adopted TSFMs are currently still in the open-source stage, we utilize these open-source base models to simulate the API deployment environment by strictly disabling gradient backpropagation. Other experimental details are provided in Appendix \ref{app:exp_details}, such as the comprehensive ORCA settings (decomposition kernel, layer size, and optimizer), the online training/evaluation data framework, and settings of baselines.

\definecolor{JapaneseLaurel}{rgb}{0,0.498,0}
\begin{table}
\centering
\small
\caption{Main experimental results comparing the vanilla zero-shot performance of five base TSFMs and their ORCA-refined counterparts. The results are averaged over three forecasting horizons ($H \in \{30, 96, 336\}$), with each horizon provided in Appendix \ref{app:horizon_breakdown}. A negative percentage indicates a reduction in MSE, meaning positive performance improvement.}
\label{tab:main_results}
\begin{tblr}{
  width = \linewidth,
  colspec = {Q[90]Q[70]Q[63]Q[70]Q[63]Q[70]Q[63]Q[70]Q[63]Q[70]Q[63]Q[70]Q[70]},
  colsep = 2pt,
  cells = {c},
  row{1} = {font=\bfseries},
  cell{1}{1} = {c=2}{0.146\linewidth},
  cell{1}{3} = {c=2}{0.134\linewidth},
  cell{1}{5} = {c=2}{0.14\linewidth},
  cell{1}{7} = {c=2}{0.14\linewidth},
  cell{1}{9} = {c=2}{0.14\linewidth},
  cell{1}{11} = {c=2}{0.14\linewidth},
  cell{2}{1} = {r=2}{font=\bfseries},
  cell{3}{4} = {fg=JapaneseLaurel},
  cell{3}{6} = {fg=JapaneseLaurel},
  cell{3}{8} = {fg=JapaneseLaurel},
  cell{3}{10} = {fg=JapaneseLaurel},
  cell{3}{12} = {fg=red},
  cell{3}{13} = {fg=JapaneseLaurel},
  cell{4}{1} = {r=2}{font=\bfseries},
  cell{5}{4} = {fg=JapaneseLaurel},
  cell{5}{6} = {fg=red},
  cell{5}{8} = {fg=JapaneseLaurel},
  cell{5}{10} = {fg=JapaneseLaurel},
  cell{5}{12} = {fg=red},
  cell{5}{13} = {fg=JapaneseLaurel},
  cell{6}{1} = {r=2}{font=\bfseries},
  cell{7}{4} = {fg=JapaneseLaurel},
  cell{7}{6} = {fg=JapaneseLaurel},
  cell{7}{8} = {fg=JapaneseLaurel},
  cell{7}{10} = {fg=JapaneseLaurel},
  cell{7}{12} = {fg=JapaneseLaurel},
  cell{7}{13} = {fg=JapaneseLaurel},
  cell{8}{1} = {r=2}{font=\bfseries},
  cell{9}{4} = {fg=JapaneseLaurel},
  cell{9}{6} = {fg=JapaneseLaurel},
  cell{9}{8} = {fg=JapaneseLaurel},
  cell{9}{10} = {fg=JapaneseLaurel},
  cell{9}{12} = {fg=JapaneseLaurel},
  cell{9}{13} = {fg=JapaneseLaurel},
  cell{10}{1} = {r=2}{font=\bfseries},
  cell{11}{4} = {fg=JapaneseLaurel},
  cell{11}{6} = {fg=red},
  cell{11}{8} = {fg=JapaneseLaurel},
  cell{11}{10} = {fg=JapaneseLaurel},
  cell{11}{12} = {fg=JapaneseLaurel},
  cell{11}{13} = {fg=JapaneseLaurel},
  cell{12}{1} = {r=2}{font=\bfseries},
  cell{13}{4} = {fg=JapaneseLaurel},
  cell{13}{6} = {fg=JapaneseLaurel},
  cell{13}{8} = {fg=JapaneseLaurel},
  cell{13}{10} = {fg=JapaneseLaurel},
  cell{13}{12} = {fg=JapaneseLaurel},
  cell{13}{13} = {fg=JapaneseLaurel},
  cell{14}{1} = {r=2}{font=\bfseries},
  cell{15}{4} = {fg=JapaneseLaurel},
  cell{15}{6} = {fg=JapaneseLaurel},
  cell{15}{8} = {fg=JapaneseLaurel},
  cell{15}{10} = {fg=JapaneseLaurel},
  cell{15}{12} = {fg=JapaneseLaurel},
  cell{15}{13} = {fg=JapaneseLaurel},
  cell{16}{1} = {r=2}{font=\bfseries},
  cell{17}{4} = {fg=JapaneseLaurel},
  cell{17}{6} = {fg=JapaneseLaurel},
  cell{17}{8} = {fg=JapaneseLaurel},
  cell{17}{10} = {fg=JapaneseLaurel},
  cell{17}{12} = {fg=JapaneseLaurel},
  cell{17}{13} = {fg=JapaneseLaurel},
  cell{18}{1} = {c=2}{0.146\linewidth,font=\bfseries},
  cell{18}{3} = {c=2}{0.134\linewidth,fg=JapaneseLaurel},
  cell{18}{5} = {c=2}{0.14\linewidth,fg=JapaneseLaurel},
  cell{18}{7} = {c=2}{0.14\linewidth,fg=JapaneseLaurel},
  cell{18}{9} = {c=2}{0.14\linewidth,fg=JapaneseLaurel},
  cell{18}{11} = {c=2}{0.14\linewidth,fg=JapaneseLaurel},
  cell{18}{13} = {fg=JapaneseLaurel,font=\bfseries},
  hline{1,19} = {-}{0.08em},
  hline{2} = {1-2}{lr},
  hline{2} = {3-13}{0.03em},
  hline{18} = {-}{0.05em},
}
Model       &         & Chronos-2 &        & Moirai-2 &         & TiRex  &         & TimesFM-2.5 &         & Sundial &         & Avg.    \\
ETTh1       & Vanilla & 0.2756    &        & 0.2653   &         & 0.2671 &         & 0.2771      &         & 0.2381  &         &         \\
            & Refined & 0.2605    & -5.5\% & 0.2579   & -2.8\%  & 0.2579 & -3.5\%  & 0.2634      & -4.9\%  & 0.2460  & 3.3\%   & -2.7\%  \\
ETTh2       & Vanilla & 0.0373    &        & 0.0350   &         & 0.0357 &         & 0.0369      &         & 0.0358  &         &         \\
            & Refined & 0.0366    & -1.8\% & 0.0355   & 1.3\%   & 0.0355 & -0.5\%  & 0.0366      & -0.9\%  & 0.0360  & 0.6\%   & -0.3\%  \\
ETTm1       & Vanilla & 0.2197    &        & 0.2538   &         & 0.2564 &         & 0.2229      &         & 0.2103  &         &         \\
            & Refined & 0.2038    & -7.2\% & 0.2164   & -14.7\% & 0.2149 & -16.2\% & 0.2053      & -7.9\%  & 0.2014  & -4.2\%  & -10.1\% \\
ETTm2       & Vanilla & 0.0245    &        & 0.0274   &         & 0.0263 &         & 0.0260      &         & 0.0248  &         &         \\
            & Refined & 0.0224    & -8.7\% & 0.0232   & -15.3\% & 0.0233 & -11.4\% & 0.0231      & -10.9\% & 0.0232  & -6.5\%  & -10.6\% \\
Exchange    & Vanilla & 0.0031    &        & 0.0033   &         & 0.0035 &         & 0.0033      &         & 0.0042  &         &         \\
            & Refined & 0.0031    & -0.0\% & 0.0034   & 2.0\%   & 0.0032 & -8.5\%  & 0.0032      & -3.1\%  & 0.0037  & -10.4\% & -4.0\%  \\
Weather     & Vanilla & 0.0477    &        & 0.0565   &         & 0.0542 &         & 0.0497      &         & 0.0415  &         &         \\
            & Refined & 0.0435    & -8.9\% & 0.0446   & -21.2\% & 0.0463 & -14.5\% & 0.0441      & -11.3\% & 0.0408  & -1.7\%  & -11.5\% \\
Electricity & Vanilla & 0.0571    &        & 0.0576   &         & 0.0549 &         & 0.0579      &         & 0.0484  &         &         \\
            & Refined & 0.0517    & -9.5\% & 0.0522   & -9.5\%  & 0.0503 & -8.5\%  & 0.0525      & -9.4\%  & 0.0453  & -6.4\%  & -8.7\%  \\
Traffic     & Vanilla & 0.2239    &        & 0.2246   &         & 0.2462 &         & 0.2219      &         & 0.2564  &         &         \\
            & Refined & 0.2202    & -1.7\% & 0.2217   & -1.3\%  & 0.2367 & -3.8\%  & 0.2201      & -0.8\%  & 0.2425  & -5.4\%  & -2.6\%  \\
Models Avg. &         & -5.4\%    &        & -7.7\%   &         & -8.4\% &         & -6.1\%      &         & -3.8\%  &         & -6.3\%  
\end{tblr}
\end{table}

\subsection{Main results}
\label{subsec:main_results}

We evaluate the overall performance of ORCA against the vanilla zero-shot TSFMs across a comprehensive matrix of 120 experimental configurations (8 datasets $\times$ 5 base models $\times$ 3 forecasting horizons). The averaged MSE results and the relative performance improvements are summarized in Table \ref{tab:main_results}. As illustrated, ORCA consistently reduces the forecasting error across the vast majority of scenarios. Across the 40 aggregated cases (8 datasets $\times$ 5 TSFMs), ORCA successfully decreases the base models' forecasting errors in 90\% of the evaluations. Furthermore, regulated by the Boltzmann Router, the performance degradation in the few cases remains bounded, with the maximum error increase limited to 3.3\%, whereas the maximum error reduction achieves up to 21.2\%.

Results across three random seeds are provided in Appendix~\ref{app:multiseed}.

\begin{figure}[h]
    \centering
    \includegraphics[width=\textwidth]{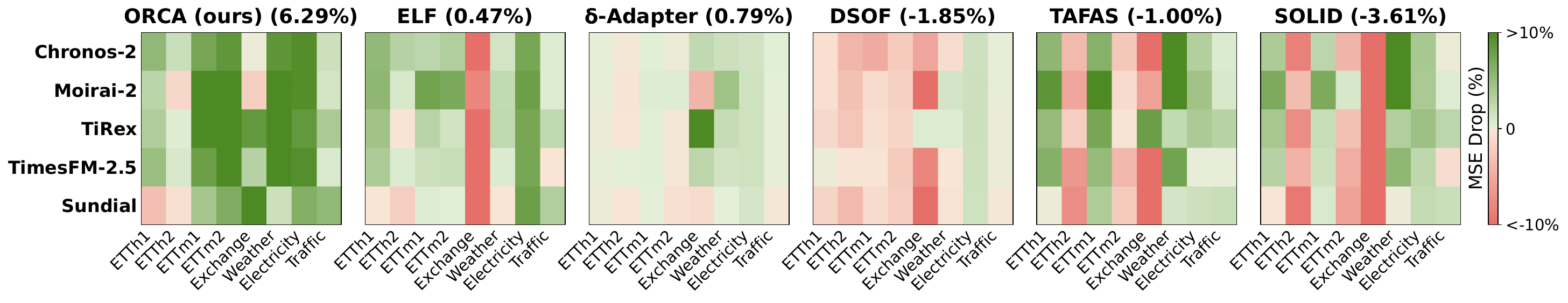}
    \caption{Heatmaps illustrating the relative MSE drop (\%) achieved by various online adaptation methods compared to the vanilla zero-shot TSFMs across 8 datasets and 5 base models. Green cells indicate a reduction in MSE (improvement), while red cells indicate an increase in MSE (degradation).}
    \label{fig:baselines}
\end{figure}

\subsection{Comparison with baselines}
\label{subsec:baselines}

To demonstrate the effectiveness of our proposed framework, we compare ORCA against comprehensive baselines, as in Figure \ref{fig:baselines}. \textbf{(1) First,} we compare SOTA (state-of-the-art) black-box online adaptation methods, including ELF \citep{leeLightweightOnlineAdaption2025} and $\delta$-Adapter (Ada-Y variant) \citep{liangForecastForecastPostProcessing2025}. \textbf{(2) However,} because general black-box online adaptation is still insufficiently explored, we have to introduce more baselines by adapting SOTA white-box and grey-box methods into the black-box paradigm (DSOF \cite{lauFastSlowStreams2024}, TAFAS \cite{kimBattlingNonstationarityTime2025a}, and SOLID \cite{chenCalibrationTimeSeriesForecasting2024}), \textbf{only as a supplement}, detailed in Appendix \ref{app:baseline_adaptation}. Importantly, they are not traditional comparisons, but rather to underscore the \textbf{research urgency and non-trivality} for more black-box TSFM adaptation methods, by proving that direct modification from white-box and grey-box methods fails. \textbf{(3) Finally,} for a fairer comparison than insufficient black-box methods and adapted white-box and grey-box methods, we added statistical baselines including Ridge Regression \citep{RidgeRegression} and error--trend--seasonality (ETS) exponential smoothing \citep{hyndmanForecastingExponentialSmoothing2008}, with settings and results detailed in Appendix \ref{app:stat_baselines}. Without proper structural design, as in ORCA or ELF, statistical methods overfit to noise; despite occasional reductions in base model errors, they can increase errors by hundreds of percent, excluded in the heatmap for readability. ETS predicts future residuals from past residuals, as it requires identical input-output sequences. Notably, Ridge Regression conditioned on context outperforms ETS, further supporting our hypothesis of learning the context of errors.

Additional forecasting comparisons are provided in Appendix~\ref{app:fullshot}.

\subsection{Quantitative analysis of different learning hypotheses}
\label{subsec:learning_context}

\begin{figure}[t]
    \centering
    \includegraphics[width=\textwidth]{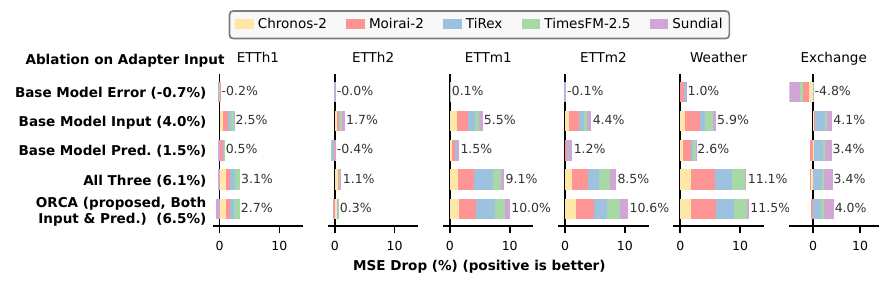}
    \caption{Ablation study on the adapter's input combinations across six datasets and five TSFMs, averaged over $H\in\{30,96,336\}$. The bars represent the MSE drop ratio (\%). Positive values indicate an improvement (error reduction) over the vanilla base model. Our proposed input (Base Model Input \& Prediction) achieves the highest average improvement of $6.5\%$.}
    \label{fig:abla_input}
\end{figure}

A question in online adaptation for black-box models is: \textit{What should the adapter learn?} To quantitatively analyze different learning hypotheses, we conduct a comprehensive ablation study on the adapter's input of the proposed ORCA. In Figure \ref{fig:abla_input}, different input configuration of the adapter is supported by different learning hypotheses. Attempting to directly learn the historical error sequence itself yields highly suboptimal results, with an average performance drop of $-0.7\%$. In several datasets, relying solely on past errors increases the forecasting error. Conversely, when we condition the adapter on either the Input or the Prediction, the performance improves. Notably, our proposed configuration, which utilizes both to obtain a comprehensive error context, achieves the optimal average improvement of $6.5\%$. Interestingly, incorporating the past error into this optimal set slightly degrades the average performance to $6.1\%$. The decline from adding past errors shows that richer inputs do not monotonically improve performance. Current inputs and base predictions are observable at prediction time, whereas the current error is only revealed with future ground truth. These findings corroborate our core insight: errors follow a conditional distribution $P(E | \boldsymbol{X}_t, \boldsymbol{Y}_t^{\mathrm{base}})$, and therefore we should learn from the context of errors.

\subsection{Ablation and sensitivity analysis}
\label{subsec:ablation_sensitivity}

\begin{figure}[b]
    \centering
    \includegraphics[width=\textwidth]{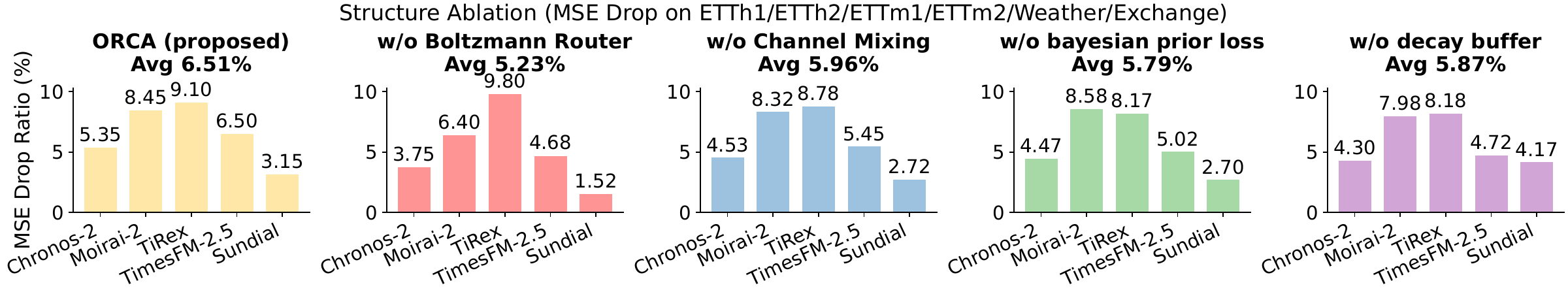}
    \caption{Structural ablation study across six datasets and five TSFMs, averaged over $H\in\{30,96,336\}$. The average MSE drop ratio falls from $6.51\%$ to $5.23\%$ without the Boltzmann Router, highlighting its critical role in harnessing online stability.}
    \label{fig:abla_structure}
\end{figure}

As depicted in Figure \ref{fig:abla_structure}, removing any of the proposed components leads to a performance decline, especially the Boltzmann Router. This validates the effectiveness of the Boltzmann routing mechanism in safely harnessing the adaptation. Without the router, the framework still retains about 80\% of the full improvement, indicating that routing complements an effective learned correction. Furthermore, the Linear Channel Mixing layers and the Bayesian prior loss also provide contributions. We also examine the role of the exponential decay FIFO buffer by testing a variant without a decay buffer, with a use-and-discard buffer (size 3000, batch size 256). It waits until 3000 samples are collected, triggers training, completely flushes all memory, and then waits to fill again. To further validate the necessity of the proposed Boltzmann routing, we tested a router variant, a hard binary router, detailed in Appendix \ref{app:router}. With the same EMA error mechanism, the binary variant only achieves a MSE reduction of 3.80\%, indicating the effectiveness of our Boltzmann router. Comparing the structural ablations with the input ablations from Section \ref{subsec:learning_context}, the wrong learning hypotheses reduce the performance more severely. This contrast proves our insight: in black-box online adaptation, \textit{what} the model learns is rather impactful.

Local regularization costs and a directed-prior comparison are examined in Appendix~\ref{app:bayesian_diagnostics}.

\begin{figure}[t]
    \centering
    \includegraphics[width=\textwidth]{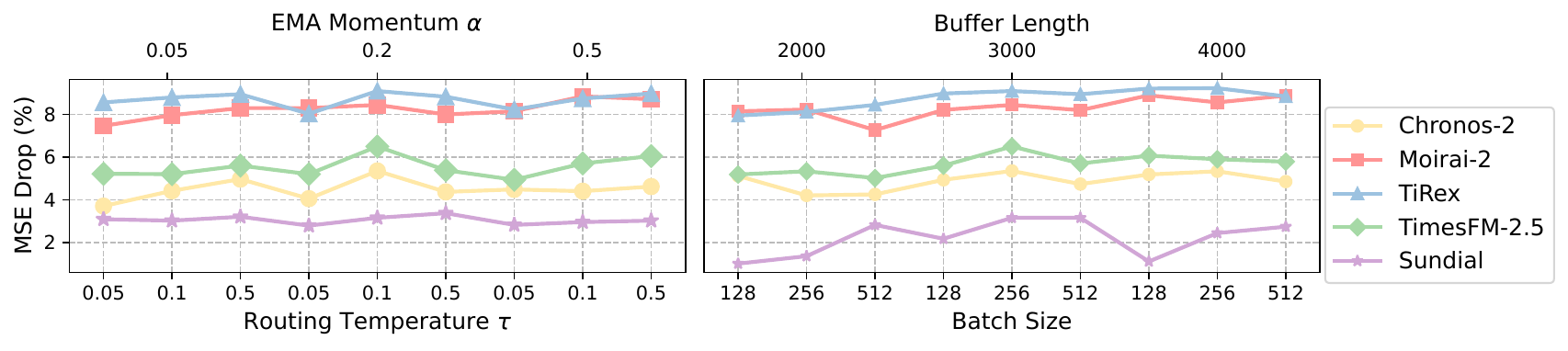}
    \caption{Hyperparameter sensitivity analysis of ORCA across different base models in 6 datasets. The relative MSE drop (\%) remains generally stable under varying routing temperatures $\tau$, EMA momentum $\alpha$, FIFO buffer lengths, and sampling batch sizes.}
    \label{fig:hyper_mse}
\end{figure}

In the ORCA framework, the hyperparameters primarily involve the Boltzmann Router and the online training configuration. For the Boltzmann Router, the routing temperature $\tau$ and the Exponential Moving Average (EMA) momentum $\alpha$ govern the confidence allocation. Guided by our theoretical analysis (detailed in Appendix \ref{app:boltzmann_sensitivity}), we set the default values to $\tau=0.1$ and $\alpha=0.2$. Empirical evaluations, as illustrated in Figure \ref{fig:hyper_mse}, demonstrate that the relative MSE drop across five base TSFMs and 6 datasets, as in Figure \ref{fig:hyper_mse}, remains stable and insensitive under varying combinations of $\tau$ and $\alpha$. Beyond the router, controlling the data instances the adapter encounters during each training cycle is crucial for balancing plasticity and stability. The two primary parameters governing this mechanism are the length of the decay FIFO buffer and the random sampling batch size. As further illustrated in Figure \ref{fig:hyper_mse}, varying the buffer length (from 2000 to 4000) and the batch size (from 128 to 512) also results in insignificant fluctuations in performance. Consequently, our default configuration of a buffer length of 3000 and a batch size of 256 was chosen directly based on device capacity.

\begin{table}[t]
\centering
\caption{Computational efficiency of ORCA on ETTh1 and Electricity datasets with Chronos-2 and Moirai-2, evaluated on a single NVIDIA B200 GPU with a forecasting horizon of $H=96$.}
\label{tab:efficiency}
\small
\begin{tblr}{
  width = \linewidth,
  colspec = {Q[106]Q[200]Q[110]Q[87]Q[163]Q[110]Q[163]},
  cells = {c},
  colsep = 2pt,
  cell{1}{1} = {r=2}{},
  cell{1}{2} = {r=2}{},
  cell{1}{3} = {c=3}{0.36\linewidth},
  cell{1}{6} = {c=2}{0.273\linewidth},
  cell{3,6}{1} = {c=7}{font=\bfseries},
  hline{1,9} = {-}{0.08em},
  hline{2} = {3-7}{lr},
  hline{3,6} = {-}{0.05em},
}
\textbf{Dataset} & {\textbf{Base Inference} \\\textbf{Time (ms per step)}} & \textbf{ORCA Inference (per step)} & & & \textbf{ORCA Training (per cycle)} & \\
 & & Time (ms) & FLOPs & GPU Usage (MB) & Time (ms) & GPU Usage (MB) \\
\underline{Chronos-2} & & & & & & \\
ETTh1 & 121.45 & 16.74 & $2.5\times10^6$ & 353.9 & 2049.2 & 463.0 \\
Electricity & 167.30 & 12.61 & $1.1\times10^8$ & 4236.3 & 2115.9 & 6216.5 \\
\underline{Moirai-2} & & & & & & \\
ETTh1 & 152.90 & 23.82 & $2.5\times10^6$ & 161.3 & 2115.7 & 270.7 \\
Electricity & 386.57 & 25.68 & $1.1\times10^8$ & 4041.6 & 3520.9 & 6017.8
\end{tblr}
\end{table}

\subsection{Performance and efficiency analysis}
\label{subsec:efficiency}

We analyze the efficiency of ORCA on two datasets: ETTh1 (7 channels) and Electricity (321 channels). We profile Chronos-2 and Moirai-2. The evaluation is conducted on a single NVIDIA B200 graphics processing unit (GPU) with a forecasting horizon of $H=96$. As shown in Table \ref{tab:efficiency}, we consider a non-overlapping evaluation setting where the online cycle training is executed once every 96 steps. Across Chronos-2 and Moirai-2, the amortized ORCA overhead is 34.65--62.36 ms per step, excluding base-model inference. Amortized overhead is the adapter inference time plus one training-cycle time divided by 96. This leaves substantial headroom for minute-level forecasting, although synchronous training can introduce a 2.05--3.52 s pause once per cycle. Asynchronous training could make second-level deployment feasible. ORCA requires $2.48\times10^6$ and $1.14\times10^8$ floating-point operations (FLOPs) per inference step on ETTh1 and Electricity, respectively. Furthermore, it is important to note that the reported base model inference time reflects a local hardware deployment. While actual closed-source TSFM API latencies may fluctuate, the client-side overhead introduced by ORCA remains lightweight.

\section{Conclusion and limitations}

\textbf{Conclusion.} We addressed the critical challenge of adapting frozen, black-box TSFMs to streaming data. We introduced ORCA, an online adaptation framework that enhances TSFM predictions without requiring access to internal model parameters. Extensive evaluations across diverse datasets and base models demonstrate that ORCA consistently reduces errors while maintaining low latency.

\textbf{Limitations.} As a post-hoc corrector, ORCA's performance relies on the base model providing reasonable forecasts. Furthermore, while we have established evaluations on standard benchmark datasets with deterministic metrics, future work will explore its deployment in more complex data featuring abrupt structural shifts and with probabilistic metrics. Local smoothing costs remain a consideration; targeted dynamic metrics and directed regularization are promising directions beyond aggregate MSE.
\newpage
{
\small
\bibliographystyle{plainnat}
\bibliography{reference}
}

\newpage
\appendix
\section*{Appendix}
\titlecontents{section}[0em]
  {\vspace{0.5em}\bfseries} 
  {\thecontentslabel\quad}
  {}
  {\titlerule*[0.8pc]{.}\contentspage}

\titlecontents{subsection}[1.5em]
  {\normalfont\small}
  {\thecontentslabel\quad}
  {}
  {\titlerule*[0.8pc]{.}\contentspage}

\titlecontents{subsubsection}[3em]
  {\itshape\footnotesize}
  {\thecontentslabel\quad}
  {}
  {\titlerule*[0.8pc]{.}\contentspage}

\startcontents[appendix] 
\printcontents[appendix]{l}{1}{\setcounter{tocdepth}{3}}
\vspace{2em}

\section{Experiment details}
\label{app:exp_details}
In addition, we split each dataset chronologically into training, validation, and testing sets following the standard 7:1:2 ratio. Since the evaluated TSFMs are zero-shot, no actual pre-training or fine-tuning of the base models is required. Our proposed ORCA adapter can seamlessly modify the base model's output immediately after a brief warmup phase (i.e., the first time the FIFO decay buffer has acquired enough data as buffer length). However, to ensure a fair and aligned comparison with existing literature and baselines, all reported metrics are strictly evaluated on the testing set portion of the data. Mean Squared Error (MSE) serves as our primary evaluation metric. Detailed formulas for the metrics are provided in this appendix.

\subsection{Online training and evaluation framework}

In the streaming time series forecasting scenario, it is critical to prevent future data leakage. Our online training and evaluation framework adheres to chronological boundaries. At time step $t$, the base model and the adapter only have access to the historical look-back window $\mathbf{X}_t$ to predict the future horizon $\mathbf{Y}_t$. The ground truth for this prediction, $\mathbf{Y}_t^{\mathrm{GT}}$, spans from $t$ to $t+H-1$. To avoid data leakage, this ground truth is not revealed to the model immediately. Instead, the framework waits until step $t+H$, when the true values of the entire horizon become fully observable. Only then is the complete snapshot paired and pushed into the FIFO replay buffer for cycle training. This mechanism ensures that the online adaptation relies solely on retrospective data, simulating real-world streaming deployments.

\subsection{Model configuration}

The complete ORCA model uses a standardized set of hyperparameters across all base models. Specifically, the Linear Adapter consists of 2 blocks with a hidden dimension of 128. The refiner context input concatenates the historical look-back $\mathbf{X}_t$ and the base model prediction $\mathbf{Y}_t^{\mathrm{base}}$. During the streaming cycle training, we maintain a FIFO replay buffer with a capacity of 3000 snapshots. At each training cycle, we draw batches of size 256. The optimizer is AdamW with a learning rate of $1 \times 10^{-4}$ and a weight decay of $1 \times 10^{-5}$. Furthermore, to strictly align with the continuous online learning paradigm and maintain cycle efficiency, the adapter undergoes a fixed number of parameter update steps during each training cycle without employing any early stopping mechanisms. To enforce sparsity, the Bayesian loss regularization weight via the L1 channel mixer is set to $1 \times 10^{-3}$. The core kernel size parameter for the moving average decomposition is dynamically determined by the prediction length $H$. Consistent with prior works, we set the kernel size to 25 when $H > 30$, and to 7 when $H \le 30$. This configuration is specifically designed to effectively capture structural temporal patterns whenever sufficient contextual information is available. When a new base model begins its streaming inference, the adapter undergoes an initial warmup phase (first buffer settled) of 50 epochs on the buffer, after which it performs 10 gradient update steps per cycle. For the Boltzmann Router, the routing temperature is set to $\tau = 0.1$, and the exponential moving average (EMA) momentum for tracking the errors is $\alpha = 0.2$. The update rule is configured to the Bayesian mode to anchor the adapter output to the historical prior.

\subsection{Evaluation metrics}

To evaluate the forecasting performance, we employ Mean Squared Error (MSE). Because different time series foundation models employ distinct internal normalization techniques that are embedded as black-box operations, their direct loss magnitudes are mathematically incompatible. To establish a fair comparison, we implement a global scalar normalization. For each dataset, we compute a single dataset-level scalar $\sigma_{\mathrm{global}}$ representing the overall scale of the time series. Both the base models' predictions and the ORCA adapter's refined predictions are first evaluated against the ground truth in the raw, unnormalized domain. Subsequently, the calculated errors are divided by this dataset-specific global scalar $\sigma_{\mathrm{global}}$ (or its squared value for MSE). The globally normalized metrics are defined as follows:
\begin{align}
    \mathrm{MSE} &= \frac{1}{\sigma_{\mathrm{global}}^2} \frac{1}{H \times D} \sum_{i=1}^H \sum_{j=1}^D (Y_{i,j}^{\mathrm{GT}} - Y_{i,j}^{\mathrm{pred}})^2 
\end{align}
Here $Y^{\mathrm{pred}}$ denotes either the base or refined forecast, with $i$ indexing forecast steps and $j$ indexing channels. This approach ensures that the improvements achieved by ORCA are consistently comparable across all combinations of datasets and base models. Furthermore, because this normalization applies a constant scalar multiplier to both the base model and the adapted model's errors, the relative MSE drop ratio remains strictly identical before and after normalization. This guarantees that the reported performance improvements are solely attributed to our adaptation framework and not an artifact of the normalization process.

\subsection{Adaptation of white-box and grey-box baselines to black-box settings}
\label{app:baseline_adaptation}

To rigorously benchmark our proposed framework, we selected state-of-the-art Test-Time Adaptation (TTA) methods originally designed for white-box or grey-box settings and adapted them to the black-box contract. This contract strictly prohibits backpropagation through the base forecaster and avoids computationally expensive re-evaluations of the base model. Below, we detail the adaptations made for each baseline to ensure a fair comparison while preserving their core mechanisms.

\textbf{TAFAS.} The original TAFAS framework introduces Periodicity-Aware Adaptation Scheduling (PAAS) alongside Input and Output Gated Calibration Modules (GCMs). It calibrates both the input context and the output predictions by continuously backpropagating through the source forecaster using partially observed ground truths. Under the black-box setting, we cannot backpropagate through the base forecaster or re-evaluate it with recalibrated inputs. Therefore, our adaptation retains only the Output GCM as a post-hoc residual calibrator. The core mechanism of the Output GCM—a variable-wise residual transformation regulated by a tanh gating mechanism initialized near zero—is fully preserved. Instead of PAAS, which relies on partial ground truth, the adapted TAFAS uses the same fully resolved ground truth training snapshots and warmup scheduling as our proposed framework, ensuring it operates strictly on observable retrospective data.

\textbf{SOLID.} The Sample-level Contextualized Adapter (SOLID) constructs a contextualized dataset for each test sample by selecting historical instances with similar periodic phases and high Euclidean similarity in their look-back windows. It then fine-tunes the top linear prediction layer of the base forecaster using this contextualized dataset. Under the black-box contract, modifying the internal prediction layer of the base forecaster is strictly forbidden. Consequently, we replace the internal prediction layer fine-tuning with a lightweight post-hoc Affine Head (comprising scale and bias parameters) applied directly to the frozen base predictions. Our adaptation fully preserves SOLID's core context-selection mechanism based on phase proximity and look-back similarity. For each test instance, we select the top-$k$ nearest historical neighbors from an online history pool. We then clone the globally trained Affine Head, perform local gradient descent adaptation using these contextualized neighbors, and apply the locally adapted head to the current prediction.

\textbf{DSOF.} The Dual-Stream Framework for Online Time Series Forecasting (DSOF) introduces a teacher-student architecture operating on two distinct data streams. The fast stream employs temporal difference (TD) learning with pseudo-labels to rapidly adapt to recent data, while the slow stream stabilizes training via experience replay (ER) using fully observed ground truth sequences. In its original formulation, both the teacher and student models are fine-tuned. To align with the black-box contract, our adaptation strictly freezes the teacher model (i.e., the base forecaster) and relies entirely on a lightweight residual student multilayer perceptron (MLP) to calibrate predictions. The dual-stream mechanism is preserved in its entirety. The slow stream utilizes the same fully resolved ground truth training snapshots from our online FIFO buffer for experience replay. Concurrently, the fast stream immediately updates the student model using pseudo-labels, which are constructed by concatenating the most recent partial ground truth observations with the frozen teacher model's predictions.

\section{Theoretical analysis}
\label{sec:theoretical_analysis}
\subsection{Context conditioned learning}
\label{app:context_conditioned}

In Section \ref{sec:methodology}, we assert that the optimal deterministic adapter should learn the conditional expectation of the errors given the context $\mathbf{C}_t = [\mathbf{X}_t, \mathbf{Y}_t^{\mathrm{base}}]$. Here, we provide the detailed mathematical derivation.

We define the expected risk $\mathcal{R}(f)$ under the Mean Squared Error (MSE) loss as the expected Frobenius norm of the difference between the true error matrix $\mathbf{E}_t$ and our adapter's prediction $f(\mathbf{C}_t)$:
\begin{equation}
    \mathcal{R}(f) = \mathbb{E}_{\mathbf{C}_t, \mathbf{E}_t} \left[ \| \mathbf{E}_t - f(\mathbf{C}_t) \|_F^2 \right]
\end{equation}
According to the law of total expectation, this risk can be decomposed by conditioning on the observable context $\mathbf{C}_t$:
\begin{equation}
    \mathcal{R}(f) = \mathbb{E}_{\mathbf{C}_t} \left[ \mathbb{E}_{\mathbf{E}_t \mid \mathbf{C}_t} \left[ \| \mathbf{E}_t - f(\mathbf{C}_t) \|_F^2 \mid \mathbf{C}_t \right] \right]
\end{equation}
To find the optimal mapping $f^*$ that minimizes the global expected risk $\mathcal{R}(f)$, it is sufficient to minimize the inner conditional expectation for every realization of the context $\mathbf{C}_t$. Let $\mathcal{J}(f(\mathbf{C}_t))$ denote this inner objective, which can be expanded over the matrix elements:
\begin{equation}
    \mathcal{J}(f(\mathbf{C}_t)) = \mathbb{E}_{\mathbf{E}_t \mid \mathbf{C}_t} \left[ \sum_{i,j} \left( E_{t,ij} - f_{ij}(\mathbf{C}_t) \right)^2 \mathrel{\Bigg|} \mathbf{C}_t \right]
\end{equation}
Since $f(\mathbf{C}_t)$ is a deterministic function of $\mathbf{C}_t$, we can take the partial derivative of $\mathcal{J}$ with respect to each output element $f_{ij}(\mathbf{C}_t)$ and set it to zero:
\begin{equation}
    \frac{\partial \mathcal{J}}{\partial f_{ij}(\mathbf{C}_t)} = -2 \, \mathbb{E}_{\mathbf{E}_t \mid \mathbf{C}_t} \left[ E_{t,ij} - f_{ij}(\mathbf{C}_t) \mid \mathbf{C}_t \right] = 0
\end{equation}
Solving this yields the optimal element-wise mapping:
\begin{equation}
    f_{ij}^*(\mathbf{C}_t) = \mathbb{E}_{\mathbf{E}_t \mid \mathbf{C}_t} \left[ E_{t,ij} \mid \mathbf{C}_t \right]
\end{equation}
Reconstructing the matrix form, we obtain the rigorously optimal mapping function:
\begin{equation}
    f^*(\mathbf{C}_t) = \mathbb{E} [\mathbf{E}_t \mid \mathbf{X}_t, \mathbf{Y}_t^{\mathrm{base}}]
\end{equation}
While it is a well-known mathematical property that minimizing MSE targets the conditional mean, this derivation serves a specific structural purpose for ORCA. It rigorously justifies our use of the conditional formulation $P(\mathbf{E}_t \mid \mathbf{X}_t, \mathbf{Y}_t^{\mathrm{base}})$ within a deterministic framework, demonstrating that adapting via a deterministic mapping $f(\mathbf{C}_t)$ is mathematically equivalent to estimating the expected value of the probabilistic error distribution. Consequently, to effectively perform this residual regression, it is naturally appropriate to condition the adapter on the concatenated context of $\mathbf{X}_t$ and $\mathbf{Y}_t^{\mathrm{base}}$.

\subsection{Predictive-space Bayesian update}
\label{app:bayesian}

In Section \ref{sec:bayesian}, we formulated our online adaptation as a predictive-space Bayesian update and substituted the exact precision ratio $\lambda_t$ with the Boltzmann routing confidence $\bar{c}_t$. This section provides the comprehensive mathematical proof and theoretical justification.

\subsubsection{Maximum a posteriori (MAP) derivation}
At step $t$, we seek to estimate the true residual $\mathbf{E}_t$ given the context $\mathbf{C}_t = [\mathbf{X}_t, \mathbf{Y}_t^{\mathrm{base}}]$, the historical memory $\mathcal{H}_{t-1}$, and the current observation $\mathbf{E}_{\mathrm{obs}, t}$. According to Bayes' theorem, the posterior distribution is:
\begin{equation}
    P(\mathbf{E}_t \mid \mathbf{C}_t, \mathbf{E}_{\mathrm{obs}, t}, \mathcal{H}_{t-1}) \propto P(\mathbf{E}_{\mathrm{obs}, t} \mid \mathbf{E}_t, \mathbf{C}_t) \cdot P(\mathbf{E}_t \mid \mathbf{C}_t, \mathcal{H}_{t-1})
\end{equation}
Assuming both the likelihood and the prior follow isotropic Gaussian distributions:
\begin{align}
    \text{Likelihood: } & P(\mathbf{E}_{\mathrm{obs}, t} \mid \mathbf{E}_t, \mathbf{C}_t) \propto \exp \left( - \frac{1}{2\sigma_{\mathrm{obs}}^2} \|\mathbf{E}_t - \mathbf{E}_{\mathrm{obs}, t}\|_F^2 \right) \\
    \text{Prior: } & P(\mathbf{E}_t \mid \mathbf{C}_t, \mathcal{H}_{t-1}) \propto \exp \left( - \frac{1}{2\sigma_{\mathrm{prior}}^2} \|\mathbf{E}_t - \mathbf{E}_{\mathrm{prior}, t}\|_F^2 \right)
\end{align}
To perform MAP estimation, we substitute the adapter's deterministic prediction $\Delta\hat{\mathbf{Y}}_t$ for $\mathbf{E}_t$ and minimize the negative logarithm of the posterior:
\begin{equation}
    -\log P \propto \frac{1}{2\sigma_{\mathrm{obs}}^2} \|\Delta\hat{\mathbf{Y}}_t - \mathbf{E}_{\mathrm{obs}, t}\|_F^2 + \frac{1}{2\sigma_{\mathrm{prior}}^2} \|\Delta\hat{\mathbf{Y}}_t - \mathbf{E}_{\mathrm{prior}, t}\|_F^2
\end{equation}
Since $\mathbf{E}_{\mathrm{obs}, t} = \mathbf{Y}_t^{\mathrm{GT}} - \mathbf{Y}_t^{\mathrm{base}}$, we have $\Delta\hat{\mathbf{Y}}_t - \mathbf{E}_{\mathrm{obs}, t} = (\mathbf{Y}_t^{\mathrm{base}} + \Delta\hat{\mathbf{Y}}_t) - \mathbf{Y}_t^{\mathrm{GT}} = \mathbf{Y}_t^{\mathrm{ada}} - \mathbf{Y}_t^{\mathrm{GT}}$. Similarly, for the prior term, $\Delta\hat{\mathbf{Y}}_t - \mathbf{E}_{\mathrm{prior}, t} = \mathbf{Y}_t^{\mathrm{ada}} - \mathbf{Y}_t^{\mathrm{prior}}$. By absorbing $2\sigma_{\mathrm{obs}}^2$ into the learning rate and dividing by the dimensions $H \times D$, we define the theoretical precision ratio $\lambda_t = \sigma_{\mathrm{obs}}^2 / \sigma_{\mathrm{prior}}^2$. This rigorously transforms the error-space optimization into the final predictive-space time series loss:
\begin{equation}
    \mathcal{L}(s_t, \theta) \propto \frac{1}{H \times D} \left( \|\mathbf{Y}_t^{\mathrm{GT}} - \mathbf{Y}_t^{\mathrm{ada}}\|_F^2 + \lambda_t \|\mathbf{Y}_t^{\mathrm{ada}} - \mathbf{Y}_t^{\mathrm{prior}}\|_F^2 \right)
\end{equation}

\subsubsection{Theoretical justification of the precision surrogate}
In online time series forecasting, the precise prior variance $\sigma_{\mathrm{prior}}^2$ is unknown and highly non-stationary due to concept drifts. While rigorous estimation of the prior precision can be mathematically possible, it often introduces unacceptable computational overhead and numerical instability for real-time streaming adaptation. To maintain both efficiency and stability, ORCA dynamically substitutes the theoretical precision ratio $\lambda_t$ with an empirical surrogate $\bar{c}_t = \mathrm{mean}(\mathbf{c}_t)$. We validate its structural optimality by solving for the minimum of the substituted loss function $\mathcal{L}^* = \|\Delta\hat{\mathbf{Y}}_t - \mathbf{E}_{\mathrm{obs}, t}\|_F^2 + \bar{c}_t \|\Delta\hat{\mathbf{Y}}_t - \mathbf{E}_{\mathrm{prior}, t}\|_F^2$.

Taking the derivative with respect to $\Delta\hat{\mathbf{Y}}_t$ and setting it to zero yields:
\begin{equation}
    \frac{\partial \mathcal{L}^*}{\partial \Delta\hat{\mathbf{Y}}_t} = 2(\Delta\hat{\mathbf{Y}}_t - \mathbf{E}_{\mathrm{obs}, t}) + 2\bar{c}_t(\Delta\hat{\mathbf{Y}}_t - \mathbf{E}_{\mathrm{prior}, t}) = 0
\end{equation}
Solving for the optimal adapter output $\Delta\hat{\mathbf{Y}}_t^*$ gives Equation~\eqref{eq:bayesian_optimal}.
Let $\mu_{\mathrm{post}}$, $\mu_{\mathrm{obs}}$, and $\mu_{\mathrm{prior}}$ denote the posterior, observation, and prior means, respectively. This formulation corresponds exactly to the analytical mean of the product of two Gaussian distributions: $\mu_{\mathrm{post}} = \frac{\sigma_{\mathrm{prior}}^2 \mu_{\mathrm{obs}} + \sigma_{\mathrm{obs}}^2 \mu_{\mathrm{prior}}}{\sigma_{\mathrm{obs}}^2 + \sigma_{\mathrm{prior}}^2}$. Setting $\sigma_{\mathrm{obs}}^2 = 1$ and surrogate variance $\tilde{\sigma}_{\mathrm{prior}}^2 = 1 / \bar{c}_t$ mathematically bridges our empirical formulation with the exact Bayesian posterior mean structure.

Furthermore, utilizing the Boltzmann routing confidence as a surrogate provides significant advantages in online stability. Unlike the theoretical precision ratio $\lambda_t \in [0, \infty)$, which can become unbounded and trigger gradient explosion under extreme distribution shifts, $\bar{c}_t \in (0, 1)$ ensures a strictly bounded loss landscape. Additionally, the exponential moving average (EMA) of absolute errors acts as a robust proxy for predictive variance, offering superior resilience against heavy-tailed outliers compared to traditional squared variance estimation. Therefore, $\bar{c}_t$ dynamically behaves as a regularized pre-conditioner that enables adaptive scaling without the burden of explicit variance tracking.

\subsection{Boltzmann routing}
\label{app:boltzmann}

\subsubsection{Regret bound of Boltzmann routing}
\label{app:boltzmann_regret}

Beyond training, the Boltzmann Router manages the inference-stage integration of $\mathbf{Y}_t^{\mathrm{base}}$ and $\mathbf{Y}_t^{\mathrm{ada}}$ into the final output $\mathbf{Y}_t^{\mathrm{comb}}$. This mechanism can be formally analyzed as an online \textit{Prediction with Expert Advice} problem \citep{freundDecisionTheoreticGeneralizationOnLine1997}.

Let $N=2$ be the number of experts, corresponding to the base model (Expert 1) and the adapter (Expert 2). At each step $t$, the router assigns a normalized weight $\mathbf{c}_t \in [0, 1]^D$ to the adapter and $(\mathbf{1} - \mathbf{c}_t)$ to the base model. To maintain plasticity in drifting environments, ORCA utilizes an Exponential Moving Average (EMA) of errors:
\begin{equation}
    \begin{cases}
        \mathbf{\hat{\varepsilon}}_{t-1}^{\mathrm{ada}} = \alpha \mathbf{e}_{t-1}^{\mathrm{ada}} + (1 - \alpha) \mathbf{\hat{\varepsilon}}_{t-2}^{\mathrm{ada}} \\
        \mathbf{\hat{\varepsilon}}_{t-1}^{\mathrm{base}} = \alpha \mathbf{e}_{t-1}^{\mathrm{base}} + (1 - \alpha) \mathbf{\hat{\varepsilon}}_{t-2}^{\mathrm{base}}
    \end{cases}
\end{equation}
where $\alpha \in (0,1)$ is the momentum coefficient. The channel-wise routing confidence vector $\mathbf{c}_t \in \mathbb{R}^D$ is then calculated via a Boltzmann softmax function with temperature $\tau$:
\begin{equation}
    \mathbf{c}_t = \frac{\exp(-\mathbf{\hat{\varepsilon}}_{t-1}^{\mathrm{ada}} / \tau)}{\exp(-\mathbf{\hat{\varepsilon}}_{t-1}^{\mathrm{base}} / \tau) + \exp(-\mathbf{\hat{\varepsilon}}_{t-1}^{\mathrm{ada}} / \tau)}
\end{equation}

To derive the regret bound, we analyze the mechanism for a single channel, dropping the channel index for brevity. Let $e_t^k \in [0, M]$ denote the instantaneous bounded error for expert $k \in \{\mathrm{base}, \mathrm{ada}\}$. Because standard error metrics (such as MSE) are convex, the loss of the combined prediction $\mathbf{Y}_t^{\mathrm{comb}} = \mathbf{c}_t \odot \mathbf{Y}_t^{\mathrm{ada}} + (\mathbf{1} - \mathbf{c}_t) \odot \mathbf{Y}_t^{\mathrm{base}}$ is upper-bounded by the expected loss under the routing distribution:
\begin{equation}
    \mathcal{L}_t(\mathbf{Y}_t^{\mathrm{comb}}) \le c_t e_t^{\mathrm{ada}} + (1 - c_t) e_t^{\mathrm{base}}
\end{equation}

By expanding the recursive EMA formulation, the smoothed error is equivalent to a discounted sum of all past errors: $\hat{\varepsilon}_{t-1}^k = \alpha \sum_{s=1}^{t-1} (1-\alpha)^{t-1-s} e_s^k$. Substituting this into the Boltzmann softmax reveals that the routing probability is exactly proportional to:
\begin{equation}
    p_t^k \propto \exp\left( - \frac{\alpha}{\tau} \sum_{s=1}^{t-1} (1-\alpha)^{t-1-s} e_s^k \right)
\end{equation}
This formulation proves that our EMA-based Boltzmann routing is mathematically equivalent to the \textit{Discounted Exponential Weights} (DEW) algorithm \citep{herbsterTrackingBestExpert1998}, operating with a discount factor $\gamma = 1 - \alpha$ and an effective learning rate $\eta = \alpha / \tau$.

Let $R_T$ denote the static regret over $T$ steps against the best single expert in hindsight. The discounting restricts the algorithm's effective memory to approximately $1/\alpha$ steps. According to the standard theoretical analysis of DEW, the regret consists of the standard Exponential Weights bound over the effective window plus a tracking penalty bias proportional to the discount rate. Using Hoeffding's lemma for bounded losses, the regret is bounded by:
\begin{equation}
    R_T = \sum_{t=1}^T \mathcal{L}_t(\mathbf{Y}_t^{\mathrm{comb}}) - \min_{k \in \{\mathrm{base}, \mathrm{ada}\}} \sum_{t=1}^T e_t^k \le \frac{\ln 2}{\eta} + \frac{\eta}{8} T M^2 + \alpha T M
\end{equation}
Substituting $\eta = \alpha / \tau$ back into the inequality, we obtain the bound for the Boltzmann Router:
\begin{equation}
    R_T \le \frac{\tau \ln 2}{\alpha} + \frac{\alpha T M^2}{8 \tau} + \alpha T M = \frac{\tau \ln 2}{\alpha} + \alpha T \left( \frac{M^2}{8 \tau} + M \right)
\end{equation}

To achieve a sublinear regret, we balance the terms with respect to $T$. By applying the arithmetic--geometric mean (AM--GM) inequality, the minimum of this upper bound is reached when the two terms are equal, yielding the optimal momentum coefficient $\alpha^*$:
\begin{equation}
    \alpha^* = \sqrt{\frac{8 \tau^2 \ln 2}{T (M^2 + 8 \tau M)}} = \mathcal{O}\left(\frac{1}{\sqrt{T}}\right)
\end{equation}

Substituting $\alpha^*$ back into the inequality, the regret bound becomes sublinear:
\begin{equation}
    R_T \le 2 \sqrt{\tau \ln 2 \cdot T \left( \frac{M^2}{8 \tau} + M \right)} = \mathcal{O}(\sqrt{T})
\end{equation}

Given $N=2$, the logarithmic term $\ln 2$ is a constant. This bound $\mathcal{O}(\sqrt{T})$ guarantees that the time-averaged regret $R_T / T$ converges to zero. In streaming scenarios where the horizon $T$ is not known, one can employ a time-varying momentum $\alpha_t \propto 1/\sqrt{t}$ to maintain this property. Consequently, if a distribution shift breaks the linear adapter ($\mathbf{e}_t^{\mathrm{ada}} \gg \mathbf{e}_t^{\mathrm{base}}$), the Boltzmann routing mechanism ensures that $\mathbf{c}_t \to \mathbf{0}$, falling back to the base TSFM.

\subsubsection{Hyperparameter sensitivity analysis of Boltzmann routing}
\label{app:boltzmann_sensitivity}

The derived regret bound explicitly reveals the theoretical trade-offs governed by the two primary hyperparameters of the Boltzmann Router: the temperature $\tau$ and the Exponential Moving Average (EMA) momentum $\alpha$. By analyzing the inequality $R_T \le \frac{\tau \ln 2}{\alpha} + \frac{\alpha T M^2}{8 \tau} + \alpha T M$, we can quantitatively evaluate the sensitivity of the adaptation mechanism.

\textbf{Sensitivity to Temperature ($\tau$):} The temperature parameter dictates the strictness of the routing distribution. In the regret bound, $\tau$ presents a clear dichotomy. A larger $\tau$ inflates the first term $\frac{\tau \ln 2}{\alpha}$, which represents the penalty of slow convergence toward the optimal expert. Physically, a high temperature leads to a more uniform mixing of $\mathbf{Y}_t^{\mathrm{base}}$ and $\mathbf{Y}_t^{\mathrm{ada}}$, providing stability but diluting the adapter's corrective potential. Conversely, a smaller $\tau$ minimizes this initialization penalty but heavily inflates the variance term $\frac{\alpha T M^2}{8 \tau}$. A low temperature forces the router to act greedily, making it highly sensitive to instantaneous noise and prone to oscillating drastically between the base model and the adapter.

\textbf{Sensitivity to EMA Momentum ($\alpha$):} The momentum parameter controls the effective memory length of the router, which is approximately proportional to $1/\alpha$. A small $\alpha$ implies a long memory, which effectively suppresses the tracking penalties $\frac{\alpha T M^2}{8 \tau}$ and $\alpha T M$, resulting in a highly stable routing trajectory that is robust to outlier errors. However, a heavily smoothed error drastically increases the first term $\frac{\tau \ln 2}{\alpha}$, causing a delayed response. If a sudden concept drift occurs, a small $\alpha$ prevents the router from rapidly shedding its historical confidence in a failing adapter. On the other hand, a large $\alpha$ allows for swift adaptation to recent shifts but exposes the router to high variance, potentially causing it to overreact to stochastic noise rather than genuine distribution changes.

Consequently, achieving optimal online adaptation requires a delicate balance between $\tau$ and $\alpha$. The mathematical boundary demonstrates that while the Boltzmann Router is fundamentally robust, adjusting these hyperparameters allows the adaptation to be perfectly tailored to the specific non-stationary dynamics and noise levels of the target streaming environment. Based on this analysis, we empirically set our default parameters to $\tau=0.1$ and $\alpha=0.2$. An EMA momentum of $\alpha=0.2$ provides a half-life of approximately three to four steps, which is perfectly suited for tracking the high-frequency concept drifts typical in streaming time series while smoothing out immediate stochastic noise. Simultaneously, a routing temperature of $\tau=0.1$ produces a sharp yet differentiable softmax distribution. This ensures that when the adapter's error significantly deviates from the base model's, the router swiftly shifts its confidence toward the superior forecaster, preventing negative optimization, while still allowing a probabilistic blend when their performances are comparable. Our numerical analysis in Figure \ref{fig:hyper_mse} further corroborates that this optimal region is quite broad, making the model practically insensitive to minor deviations around these default values.

\section{Supplementary experiment results}

\subsection{A router variant compared to the Boltzmann router}
\label{app:router}

\begin{table}
\centering
\caption{Ablation study on the routing mechanism: Relative MSE drop (\%) using a naive Hard Router instead of the proposed Boltzmann Router ($H=96$).}
\label{tab:hard_routing_ablation}
\small
\begin{tblr}{
  width = \linewidth,
  colspec = {Q[152]Q[138]Q[115]Q[108]Q[146]Q[102]Q[171]},
  cells = {c},
  row{1} = {font=\bfseries},
  cell{2}{1} = {font=\bfseries},
  cell{3}{1} = {font=\bfseries},
  cell{4}{1} = {font=\bfseries},
  cell{5}{1} = {font=\bfseries},
  cell{6}{1} = {font=\bfseries},
  cell{7}{1} = {font=\bfseries},
  cell{8}{1} = {font=\bfseries},
  hline{1,9} = {-}{0.08em},
  hline{2,8} = {-}{0.05em},
}
Model       & Chronos-2 & Moirai-2 & TiRex    & TimesFM-2.5 & Sundial & Datasets Avg. \\
ETTh1       & -4.40\%   & -2.00\%  & 0.00\%   & 0.70\%    & 3.60\%  & -0.40\%       \\
ETTh2       & -1.80\%   & 0.60\%   & 1.80\%   & -0.60\%   & -0.40\% & -0.10\%       \\
ETTm1       & -4.70\%   & -11.30\% & -11.60\% & -3.60\%   & 0.70\%  & -6.10\%       \\
ETTm2       & -1.40\%   & -8.30\%  & -11.40\% & -8.80\%   & -3.40\% & -6.70\%       \\
Exchange    & -2.70\%   & 1.00\%   & -7.20\%  & -2.20\%   & -9.50\% & -4.10\%       \\
Weather     & -2.30\%   & -12.90\% & -7.80\%  & -5.50\%   & 2.30\%  & -5.20\%       \\
Models Avg. & -2.90\%   & -5.50\%  & -6.10\%  & -3.30\%   & -1.10\% & -3.80\%       
\end{tblr}
\end{table}

To further validate the necessity of the proposed Boltzmann routing mechanism, we conduct an additional ablation study by replacing it with a naive Hard Router. Under the identical experimental configuration with a forecasting horizon of $H=96$, the Hard Router retains the exact same Exponential Moving Average (EMA) tracking rules and hyperparameters, but strictly outputs either the base model's prediction or the adapted prediction based solely on whichever has the lower historical smoothed error. As demonstrated in Table \ref{tab:hard_routing_ablation}, this rigid binary selection strategy yields an overall average MSE reduction of only 3.80\%. This performance is markedly inferior to the results achieved by our proposed Boltzmann Router. The comparison highlights that a hard switching mechanism is overly sensitive to instantaneous noise and local distribution shifts. In contrast, the Boltzmann Router provides a soft, probabilistic mixing mechanism that more safely and effectively harnesses the corrective potential of the online adapter.

\subsection{Horizon-wise performance breakdown}
\label{app:horizon_breakdown}

In this section, we present the detailed forecasting performance of ORCA and the zero-shot base models broken down by individual forecasting horizons: $H=30$ (Table \ref{tab:h30}), $H=96$ (Table \ref{tab:h96}), and $H=336$ (Table \ref{tab:h336}). In the tables of the appendix, the following abbreviations are used: Van. for Vanilla, Ref. for Refined, Exc. for Exchange, Elc. for Electricity, and Wea. for Weather.

As demonstrated across the three tables, ORCA achieves consistent and robust MSE reductions regardless of the forecasting length. Importantly, there is no pronounced bias or trend indicating that the adaptation is disproportionately effective for only short or long horizons. This uniform stability can be directly attributed to our online cycle training scheme. Since the training cadence is dynamically linked to the horizon (i.e., triggered every $H$ steps when the full ground truth becomes observable), the learning regime scales naturally. Short-horizon forecasts trigger frequent, rapid updates to capture fast-evolving concept drifts, whereas long-horizon forecasts accumulate broader contexts before executing more comprehensive updates. Consequently, the adapter maintains its plasticity and calibration capacity optimally tailored to the intrinsic frequency of the targeted horizon.

\begin{table}[h]
\centering
\caption{Performance breakdown for forecasting horizon $H=30$.}
\label{tab:h30}
\small
\begin{tblr}{
  width = \linewidth,
  colspec = {Q[70]Q[40]Q[63]Q[70]Q[63]Q[70]Q[63]Q[75]Q[63]Q[70]Q[63]Q[68]Q[68]},
  cells = {c},
  colsep = 2pt,
  row{1} = {font=\bfseries},
  cell{1}{1} = {c=2}{0.141\linewidth},
  cell{1}{3} = {c=2}{0.143\linewidth},
  cell{1}{5} = {c=2}{0.143\linewidth},
  cell{1}{7} = {c=2}{0.141\linewidth},
  cell{1}{9} = {c=2}{0.143\linewidth},
  cell{1}{11} = {c=2}{0.139\linewidth},
  cell{2}{1} = {r=2}{font=\bfseries},
  cell{4}{1} = {r=2}{font=\bfseries},
  cell{6}{1} = {r=2}{font=\bfseries},
  cell{8}{1} = {r=2}{font=\bfseries},
  cell{10}{1} = {r=2}{font=\bfseries},
  cell{12}{1} = {r=2}{font=\bfseries},
  cell{14}{1} = {r=2}{font=\bfseries},
  cell{16}{1} = {r=2}{font=\bfseries},
  cell{18}{1} = {c=2}{0.141\linewidth,font=\bfseries},
  cell{18}{3} = {c=2}{0.143\linewidth},
  cell{18}{5} = {c=2}{0.143\linewidth},
  cell{18}{7} = {c=2}{0.141\linewidth},
  cell{18}{9} = {c=2}{0.143\linewidth},
  cell{18}{11} = {c=2}{0.139\linewidth},
  hline{1,19} = {-}{0.08em},
  hline{2} = {1-2}{lr},
  hline{2} = {3-13}{0.03em},
  hline{18} = {-}{0.05em},
}
Model        &         & Chronos-2  &          & Moirai-2  &          & TiRex   &          & TimesFM-2.5  &          & Sundial  &         & Avg.    \\
ETTh1        & Van. & 0.2086     &          & 0.21      &          & 0.2099  &          & 0.2201     &          & 0.1903   &         &         \\
             & Ref. & 0.2062     & -1.20\%  & 0.2066    & -1.60\%  & 0.2084  & -0.70\%  & 0.2143     & -2.60\%  & 0.1979   & 4.00\%  & -0.40\% \\
ETTh2        & Van & 0.0228     &          & 0.022     &          & 0.0222  &          & 0.0233     &          & 0.022    &         &         \\
             & Ref. & 0.0225     & -1.30\%  & 0.022     & 0.00\%   & 0.0223  & 0.50\%   & 0.023      & -1.30\%  & 0.0221   & 0.50\%  & -0.30\% \\
ETTm1        & Van & 0.1365     &          & 0.1675    &          & 0.1696  &          & 0.1445     &          & 0.1371   &         &         \\
             & Ref. & 0.1325     & -2.90\%  & 0.1495    & -10.70\% & 0.1509  & -11.00\% & 0.138      & -4.50\%  & 0.1344   & -2.00\% & -6.20\% \\
ETTm2        & Van & 0.0131     &          & 0.0143    &          & 0.0141  &          & 0.0141     &          & 0.0134   &         &         \\
             & Ref. & 0.0127     & -3.10\%  & 0.0132    & -7.70\%  & 0.0132  & -6.40\%  & 0.0133     & -5.70\%  & 0.013    & -3.00\% & -5.20\% \\
Exc.     & Van & 0.0007     &          & 0.0007    &          & 0.0007  &          & 0.0007     &          & 0.0008   &         &         \\
             & Ref. & 0.0007     & 0.00\%   & 0.0007    & 0.00\%   & 0.0007  & 0.00\%   & 0.0007     & 0.00\%   & 0.0008   & 0.00\%  & 0.00\%  \\
Wea.      & Van & 0.039      &          & 0.0433    &          & 0.0448  &          & 0.0449     &          & 0.0353   &         &         \\
             & Ref. & 0.0365     & -6.40\%  & 0.0376    & -13.20\% & 0.0395  & -11.80\% & 0.0391     & -12.90\% & 0.0351   & -0.60\% & -9.00\% \\
Elc.  & Van & 0.0457     &          & 0.0469    &          & 0.0444  &          & 0.0475     &          & 0.0382   &         &         \\
             & Ref. & 0.041      & -10.30\% & 0.0421    & -10.30\% & 0.04    & -9.80\%  & 0.0428     & -9.90\%  & 0.0349   & -8.60\% & -9.80\% \\
Traffic      & Van & 0.1982     &          & 0.2005    &          & 0.2206  &          & 0.2038     &          & 0.2294   &         &         \\
             & Ref. & 0.1947     & -1.70\%  & 0.1975    & -1.50\%  & 0.2112  & -4.30\%  & 0.2011     & -1.30\%  & 0.2162   & -5.70\% & -2.90\% \\
Models Avg.  &         & -3.40\%    &          & -5.60\%   &          & -5.40\% &          & -4.80\%    &          & -1.90\%  &         & -4.20\% 
\end{tblr}
\end{table}

\begin{table}[h]
\centering
\caption{Performance breakdown for forecasting horizon $H=96$.}
\label{tab:h96}
\small
\begin{tblr}{
  width = \linewidth,
  colspec = {Q[70]Q[40]Q[63]Q[70]Q[63]Q[70]Q[63]Q[70]Q[63]Q[70]Q[63]Q[70]Q[70]},
  cells = {c},
  colsep = 2pt,
  row{1} = {font=\bfseries},
  cell{1}{1} = {c=2}{0.117\linewidth},
  cell{1}{3} = {c=2}{0.138\linewidth},
  cell{1}{5} = {c=2}{0.144\linewidth},
  cell{1}{7} = {c=2}{0.142\linewidth},
  cell{1}{9} = {c=2}{0.142\linewidth},
  cell{1}{11} = {c=2}{0.138\linewidth},
  cell{2}{1} = {r=2}{font=\bfseries},
  cell{4}{1} = {r=2}{font=\bfseries},
  cell{6}{1} = {r=2}{font=\bfseries},
  cell{8}{1} = {r=2}{font=\bfseries},
  cell{10}{1} = {r=2}{font=\bfseries},
  cell{12}{1} = {r=2}{font=\bfseries},
  cell{14}{1} = {r=2}{font=\bfseries},
  cell{16}{1} = {r=2}{font=\bfseries},
  cell{18}{1} = {c=2}{0.117\linewidth,font=\bfseries},
  hline{1,19} = {-}{0.08em},
  hline{2} = {1-2}{lr},
  hline{2} = {3-13}{0.03em},
  hline{18} = {-}{0.05em},
}
Model        &         & Chronos-2  &         & Moirai-2  &          & TiRex   &          & TimesFM-2.5  &          & Sundial  &          & Avg.     \\
ETTh1        & Van. & 0.2942     &         & 0.2789    &          & 0.2777  &          & 0.2913     &          & 0.2467   &          &          \\
             & Ref. & 0.2731     & -7.20\% & 0.2703    & -3.10\%  & 0.2684  & -3.30\%  & 0.2741     & -5.90\%  & 0.2548   & 3.30\%   & -3.20\%  \\
ETTh2        & Van. & 0.0373     &         & 0.0345    &          & 0.0351  &          & 0.0368     &          & 0.0347   &          &          \\
             & Ref. & 0.0363     & -2.70\% & 0.0345    & 0.00\%   & 0.0351  & 0.00\%   & 0.0358     & -2.70\%  & 0.035    & 0.90\%   & -0.90\%  \\
ETTm1        & Van. & 0.2123     &         & 0.2428    &          & 0.2461  &          & 0.2132     &          & 0.2051   &          &          \\
             & Ref. & 0.1982     & -6.60\% & 0.207     & -14.70\% & 0.2096  & -14.80\% & 0.1987     & -6.80\%  & 0.1957   & -4.60\%  & -9.50\%  \\
ETTm2        & Van. & 0.0224     &         & 0.025     &          & 0.0246  &          & 0.0242     &          & 0.0229   &          &          \\
             & Ref. & 0.0209     & -6.70\% & 0.0217    & -13.20\% & 0.0216  & -12.20\% & 0.0217     & -10.30\% & 0.0214   & -6.60\%  & -9.80\%  \\
Exc.     & Van. & 0.0019     &         & 0.0022    &          & 0.0021  &          & 0.0021     &          & 0.0027   &          &          \\
             & Ref. & 0.002      & 5.30\%  & 0.0023    & 4.50\%   & 0.0019  & -9.50\%  & 0.0021     & 0.00\%   & 0.0024   & -11.10\% & -2.20\%  \\
Wea.      & Van. & 0.0486     &         & 0.0603    &          & 0.0556  &          & 0.0522     &          & 0.042    &          &          \\
             & Ref. & 0.0445     & -8.40\% & 0.0469    & -22.20\% & 0.0479  & -13.80\% & 0.0459     & -12.10\% & 0.0417   & -0.70\%  & -11.50\% \\
Elc.  & Van. & 0.0564     &         & 0.0569    &          & 0.054   &          & 0.0575     &          & 0.0479   &          &          \\
             & Ref. & 0.0511     & -9.50\% & 0.0519    & -8.70\%  & 0.0497  & -8.00\%  & 0.0523     & -9.10\%  & 0.0448   & -6.40\%  & -8.30\%  \\
Traffic      & Van. & 0.2255     &         & 0.2273    &          & 0.2512  &          & 0.2257     &          & 0.2609   &          &          \\
             & Ref. & 0.2224     & -1.30\% & 0.2255    & -0.80\%  & 0.2423  & -3.50\%  & 0.2248     & -0.40\%  & 0.247    & -5.30\%  & -2.30\%  \\
Models Avg.  &         & -4.60\%    &         & -7.30\%   &          & -8.20\% &          & -5.90\%    &          & -3.80\%  &          & -6.00\%  
\end{tblr}
\end{table}

\begin{table}[h]
\caption{Performance breakdown for forecasting horizon $H=336$.}
\label{tab:h336}
\centering
\small
\begin{tblr}{
  width = \linewidth,
  colspec = {Q[70]Q[40]Q[63]Q[70]Q[63]Q[70]Q[63]Q[70]Q[63]Q[70]Q[63]Q[70]Q[70]},
  colsep = 2pt,
  cells = {c},
  row{1} = {font=\bfseries},
  cell{1}{1} = {c=2}{0.123\linewidth},
  cell{1}{3} = {c=2}{0.144\linewidth},
  cell{1}{5} = {c=2}{0.148\linewidth},
  cell{1}{7} = {c=2}{0.144\linewidth},
  cell{1}{9} = {c=2}{0.146\linewidth},
  cell{1}{11} = {c=2}{0.142\linewidth},
  cell{2}{1} = {r=2}{font=\bfseries},
  cell{4}{1} = {r=2}{font=\bfseries},
  cell{6}{1} = {r=2}{font=\bfseries},
  cell{8}{1} = {r=2}{font=\bfseries},
  cell{10}{1} = {r=2}{font=\bfseries},
  cell{12}{1} = {r=2}{font=\bfseries},
  cell{14}{1} = {r=2}{font=\bfseries},
  cell{16}{1} = {r=2}{font=\bfseries},
  cell{18}{1} = {c=2}{0.123\linewidth,font=\bfseries},
  hline{1,19} = {-}{0.08em},
  hline{2} = {1-2}{lr},
  hline{2} = {3-13}{0.03em},
  hline{18} = {-}{0.05em},
}
Model        &         & Chronos-2  &          & Moirai-2  &          & TiRex   &          & TimesFM-2.5  &          & Sundial  &          & Avg.     \\
ETTh1        & Van. & 0.3239     &          & 0.3069    &          & 0.3138  &          & 0.32       &          & 0.2774   &          &          \\
             & Ref. & 0.3023     & -6.70\%  & 0.2969    & -3.30\%  & 0.2968  & -5.40\%  & 0.3019     & -5.70\%  & 0.2854   & 2.90\%   & -3.60\%  \\
ETTh2        & Van. & 0.0518     &          & 0.0485    &          & 0.0498  &          & 0.0507     &          & 0.0508   &          &          \\
             & Ref. & 0.0511     & -1.40\%  & 0.0499    & 2.90\%   & 0.0492  & -1.20\%  & 0.051      & 0.60\%   & 0.051    & 0.40\%   & 0.30\%   \\
ETTm1        & Van. & 0.3103     &          & 0.351     &          & 0.3536  &          & 0.311      &          & 0.2887   &          &          \\
             & Ref. & 0.2808     & -9.50\%  & 0.2927    & -16.60\% & 0.2841  & -19.70\% & 0.2793     & -10.20\% & 0.2741   & -5.10\%  & -12.20\% \\
ETTm2        & Van. & 0.038      &          & 0.0429    &          & 0.0401  &          & 0.0396     &          & 0.038    &          &          \\
             & Ref. & 0.0335     & -11.80\% & 0.0347    & -19.10\% & 0.035   & -12.70\% & 0.0344     & -13.10\% & 0.0351   & -7.60\%  & -12.90\% \\
Exc.     & Van. & 0.0067     &          & 0.0071    &          & 0.0078  &          & 0.007      &          & 0.009    &          &          \\
             & Ref. & 0.0066     & -1.50\%  & 0.0072    & 1.40\%   & 0.0071  & -9.00\%  & 0.0067     & -4.30\%  & 0.008    & -11.10\% & -4.90\%  \\
Wea.      & Van. & 0.0556     &          & 0.066     &          & 0.0622  &          & 0.0521     &          & 0.0473   &          &          \\
             & Ref. & 0.0494     & -11.20\% & 0.0492    & -25.50\% & 0.0516  & -17.00\% & 0.0474     & -9.00\%  & 0.0457   & -3.40\%  & -13.20\% \\
Elc.  & Van. & 0.0693     &          & 0.0692    &          & 0.0664  &          & 0.0687     &          & 0.0593   &          &          \\
             & Ref. & 0.0631     & -8.90\%  & 0.0625    & -9.70\%  & 0.061   & -8.10\%  & 0.0624     & -9.20\%  & 0.0563   & -5.10\%  & -8.20\%  \\
Traffic      & Van. & 0.2479     &          & 0.246     &          & 0.2666  &          & 0.2361     &          & 0.2789   &          &          \\
             & Ref. & 0.2433     & -1.90\%  & 0.2421    & -1.60\%  & 0.2566  & -3.80\%  & 0.2346     & -0.60\%  & 0.2642   & -5.30\%  & -2.60\%  \\
Models Avg.  &         & -6.60\%    &          & -8.90\%   &          & -9.60\% &          & -6.40\%    &          & -4.30\%  &          & -7.20\%  
\end{tblr}
\end{table}

\subsection{Statistical baselines: setting and results}
\label{app:stat_baselines}

\paragraph{ETS.}
The ETS adapter is a lightweight channel-wise Holt-style smoother that operates on the residual sequence, rather than on the input-output context, because the classic ETS often requires that its input and output be identical sequences. At each update, it first forms the residual by subtracting the median base forecast from the ground truth, then clips extreme impulses with a residual scale factor, and updates a level-trend state with smoothing coefficients. In our final configuration, the effective settings are level and trend smoothing coefficients $\alpha=0.3$ and $\beta=0.03$, damping factor $0.98$, maximum residual history length $12000$, gain sensitivity $1.0$, residual clipping scale $3.0$, minimum history for gating $5$, and warm-up steps $3$. The prediction stage extrapolates the residual state forward and adds it back to the base forecast, with a stability gate computed from recent residual variability. This design is intentionally simple and fast, but it also assumes that the residual process is approximately homogeneous over time. Table~\ref{tab:ets_results} shows that this assumption is too restrictive for black-box online adaptation: ETS is consistently weak on average, with a dataset-averaged score of $126.30\%$, and it degrades substantially on ETTh2, ETTm1, ETTm2, and Exchange. These results suggest that modeling residuals in isolation is not sufficient when the error dynamics are nonstationary.

\paragraph{Ridge Regression.}
The Ridge adapter follows a different philosophy: instead of modeling the residual sequence alone, it learns a correction from the concatenation of the past input window and the base model output along the feature dimension. In other words, the target is still the residual correction, but the predictor is conditioned on the XY context, which makes it a direct implementation of our learning hypothesis that adaptation should learn the context of errors rather than the residual process in isolation. The final implementation uses closed-form ridge regression with ridge penalty $\lambda=10^{-3}$, collects $512$ training windows before the first fit, refits once every horizon cycle, and keeps the same lightweight robustness mechanism as a safeguard, including residual-history gating, clipping, and warm-up control. Concretely, the runtime defaults are a maximum history length of $2048$, gain sensitivity $1.0$, residual clipping scale $3.0$, minimum gating history $5$, and warm-up steps $10$. Table~\ref{tab:ridge_results} shows that this XY-conditioned formulation is much stronger than ETS: Ridge reduces the dataset-averaged score to $68.78\%$, which is far better than ETS's $126.30\%$, and it achieves especially strong improvements on ETTh1 to ETTm2, where several backbones even obtain negative relative scores. The remaining failures on Exchange and Weather indicate that a purely linear model is still limited, but the overall pattern strongly supports the claim that conditioning on the main model's XY-aligned context is more informative than predicting residuals from residuals alone. We assume that the successful error reduction on the ETT series comes from the periodic physical nature of electricity transformers. When it comes to more complicated systems like Exchange (based on society and economics) and Weather (a complex natural system), the Ridge regression is incapable.

\begin{table}
\centering
\caption{ETS results for residual-to-residual online adaptation across backbones and datasets. The results are averaged over three forecasting horizons ($H \in \{30, 96, 336\}$). A negative percentage indicates a reduction in MSE, meaning positive performance improvement.}
\label{tab:ets_results}
\small
\begin{tblr}{
  width = \linewidth,
  colspec = {Q[148]Q[137]Q[113]Q[110]Q[144]Q[110]Q[169]},
  cells = {c},
  row{1} = {font=\bfseries},
  cell{2}{1} = {font=\bfseries},
  cell{3}{1} = {font=\bfseries},
  cell{4}{1} = {font=\bfseries},
  cell{5}{1} = {font=\bfseries},
  cell{6}{1} = {font=\bfseries},
  cell{7}{1} = {font=\bfseries},
  cell{8}{1} = {font=\bfseries},
  hline{1,9} = {-}{0.08em},
  hline{2,8} = {-}{0.05em},
}
Model       & Chronos-2 & Moirai-2 & TiRex    & TimesFM-2.5 & Sundial  & Datasets Avg. \\
ETTh1       & 109.30\%  & 100.60\% & 104.60\% & 99.60\%   & 97.20\%  & 102.26\%      \\
ETTh2       & 138.60\%  & 135.60\% & 112.60\% & 120.60\%  & 94.30\%  & 120.34\%      \\
ETTm1       & 156.60\%  & 156.80\% & 148.60\% & 159.10\%  & 142.40\% & 152.70\%      \\
ETTm2       & 174.70\%  & 174.60\% & 145.70\% & 156.00\%  & 122.10\% & 154.62\%      \\
Exchange    & 169.90\%  & 201.00\% & 142.50\% & 167.30\%  & 65.60\%  & 149.26\%      \\
Weather     & 95.50\%   & 78.90\%  & 74.70\%  & 69.10\%   & 74.80\%  & 78.60\%       \\
Models Avg. & 140.77\%  & 141.25\% & 121.45\% & 128.62\%  & 99.40\%  & 126.30\%      
\end{tblr}
\end{table}

\begin{table}
\centering
\caption{Ridge regression results for XY-conditioned residual correction across backbones and datasets. The results are averaged over three forecasting horizons ($H \in \{30, 96, 336\}$). A negative percentage indicates a reduction in MSE, meaning positive performance improvement.}
\label{tab:ridge_results}
\small
\begin{tblr}{
  width = \linewidth,
  colspec = {Q[148]Q[135]Q[113]Q[112]Q[144]Q[113]Q[167]},
  cells = {c},
  row{1} = {font=\bfseries},
  cell{2}{1} = {font=\bfseries},
  cell{3}{1} = {font=\bfseries},
  cell{4}{1} = {font=\bfseries},
  cell{5}{1} = {font=\bfseries},
  cell{6}{1} = {font=\bfseries},
  cell{7}{1} = {font=\bfseries},
  cell{8}{1} = {font=\bfseries},
  hline{1,9} = {-}{0.08em},
  hline{2,8} = {-}{0.05em},
}
Model       & Chronos-2 & Moirai-2 & TiRex    & TimesFM-2.5 & Sundial  & Datasets Avg. \\
ETTh1       & -3.00\%   & -4.40\%  & -2.70\%  & -2.00\%   & -0.10\%  & -2.44\%       \\
ETTh2       & -3.70\%   & -3.20\%  & -1.80\%  & -4.20\%   & 0.60\%   & -2.46\%       \\
ETTm1       & -4.70\%   & -7.80\%  & -5.80\%  & -3.50\%   & -2.90\%  & -4.94\%       \\
ETTm2       & -5.70\%   & -9.70\%  & -6.70\%  & -6.00\%   & -2.70\%  & -6.16\%       \\
Exc.        & 137.80\%  & 122.00\% & 62.00\%  & 87.90\%   & 162.70\% & 114.48\%      \\
Wea.        & 41.30\%   & 371.30\% & 573.40\% & 382.60\%  & 202.30\% & 314.18\%      \\
Models Avg. & 27.00\%   & 78.03\%  & 103.07\% & 75.80\%   & 59.98\%  & 68.78\%       
\end{tblr}
\end{table}

\clearpage
\section{Additional empirical analyses}

\subsection{Multi-seed reliability}
\label{app:multiseed}
Table~\ref{tab:multiseed} summarizes seeds 42, 43, and 44 over three horizons, eight datasets, and five TSFMs (360 runs). After averaging horizons, we report the mean and standard deviation (SD) of relative MSE changes across seeds; negative means indicate improvements, and SD is measured in percentage points. Avg. entries average cell means and SDs separately. Variability is small except for larger fluctuations on Exchange.

\begin{table}[htbp]
\centering
\caption{Relative MSE change across three seeds, expressed as $(\mathrm{mean}\pm\mathrm{SD})\%$. All entries use percentage units.}
\label{tab:multiseed}
\footnotesize
\setlength{\tabcolsep}{2pt}
\begin{tabular}{@{}lcccccc@{}}
\toprule
Dataset & Chronos-2 & Moirai-2 & TiRex & TimesFM-2.5 & Sundial & Avg. \\
\midrule
ETTh1 & $(-5.5\pm0.05)$ & $(-2.8\pm0.04)$ & $(-3.5\pm0.05)$ & $(-4.9\pm0.06)$ & $(3.3\pm0.12)$ & $(-2.68\pm0.06)$ \\
ETTh2 & $(-1.8\pm0.10)$ & $(1.1\pm0.21)$ & $(-0.5\pm0.11)$ & $(-1.1\pm0.36)$ & $(0.6\pm0.07)$ & $(-0.34\pm0.17)$ \\
ETTm1 & $(-7.2\pm0.06)$ & $(-14.6\pm0.18)$ & $(-16.2\pm0.05)$ & $(-7.9\pm0.06)$ & $(-4.1\pm0.32)$ & $(-10.00\pm0.13)$ \\
ETTm2 & $(-8.6\pm0.25)$ & $(-15.4\pm0.10)$ & $(-11.3\pm0.11)$ & $(-10.9\pm0.15)$ & $(-6.5\pm0.00)$ & $(-10.54\pm0.12)$ \\
Exchange & $(-0.0\pm0.00)$ & $(1.5\pm1.46)$ & $(-9.1\pm0.99)$ & $(-4.2\pm1.42)$ & $(-9.9\pm0.57)$ & $(-4.34\pm0.89)$ \\
Weather & $(-8.9\pm0.00)$ & $(-21.2\pm0.05)$ & $(-14.5\pm0.00)$ & $(-11.3\pm0.00)$ & $(-1.7\pm0.00)$ & $(-11.52\pm0.01)$ \\
Electricity & $(-9.5\pm0.00)$ & $(-9.5\pm0.00)$ & $(-8.5\pm0.05)$ & $(-9.4\pm0.00)$ & $(-6.4\pm0.05)$ & $(-8.66\pm0.02)$ \\
Traffic & $(-1.7\pm0.00)$ & $(-1.3\pm0.00)$ & $(-3.8\pm0.05)$ & $(-0.8\pm0.00)$ & $(-5.4\pm0.00)$ & $(-2.60\pm0.01)$ \\
\midrule
Models Avg. & $(-5.40\pm0.06)$ & $(-7.78\pm0.26)$ & $(-8.43\pm0.18)$ & $(-6.31\pm0.26)$ & $(-3.76\pm0.14)$ & $(-6.34\pm0.18)$ \\
\bottomrule
\end{tabular}
\end{table}

\subsection{Full-shot forecasting baselines}
\label{app:fullshot}
DLinear~\citep{zengAreTransformersEffective2023b}, PatchTST~\citep{nieTimeSeriesWorth2022a}, and iTransformer~\citep{liuITransformerInvertedTransformers2023} use model configurations and training hyperparameters from the Time Series Library (TSLib)~\citep{thuml2023tslib}. We train and validate chronologically with context length 96, seed 42, and reversible instance normalization (RevIN); $H=30$ uses the $H=96$ hyperparameters. Testing follows ORCA's windows, normalization, and metrics. Table~\ref{tab:fullshot} reports the results. Their overall MSEs are 0.1509, 0.1268, and 0.1246, respectively; the five TSFMs average 0.1128 (range 0.1074--0.1180), or 0.1063 (0.1049--0.1085) with ORCA.

\begin{table}[htbp]
\centering
\caption{Full-shot baseline MSE (seed 42). Exc., Wea., and Elc. denote Exchange, Weather, and Electricity. Horizon averages are computed before rounding.}
\label{tab:fullshot}
\small
\setlength{\tabcolsep}{4pt}
\begin{tabular}{@{}lrrrrrrrr@{}}
\toprule
Model & ETTh1 & ETTh2 & ETTm1 & ETTm2 & Exc. & Wea. & Elc. & Traffic \\
\midrule
\multicolumn{9}{c}{$H=30$} \\
DLinear & 0.231 & 0.024 & 0.164 & 0.016 & 0.001 & 0.089 & 0.088 & 0.447 \\
PatchTST & 0.205 & 0.025 & 0.145 & 0.014 & 0.001 & 0.088 & 0.093 & 0.255 \\
iTransformer & 0.211 & 0.025 & 0.159 & 0.015 & 0.001 & 0.080 & 0.084 & 0.220 \\
\midrule
\multicolumn{9}{c}{$H=96$} \\
DLinear & 0.296 & 0.036 & 0.220 & 0.023 & 0.002 & 0.078 & 0.097 & 0.450 \\
PatchTST & 0.265 & 0.038 & 0.208 & 0.023 & 0.002 & 0.076 & 0.126 & 0.285 \\
iTransformer & 0.276 & 0.038 & 0.225 & 0.025 & 0.002 & 0.077 & 0.101 & 0.251 \\
\midrule
\multicolumn{9}{c}{$H=336$} \\
DLinear & 0.333 & 0.052 & 0.336 & 0.037 & 0.006 & 0.080 & 0.099 & 0.415 \\
PatchTST & 0.302 & 0.053 & 0.303 & 0.037 & 0.006 & 0.079 & 0.117 & 0.298 \\
iTransformer & 0.322 & 0.054 & 0.320 & 0.040 & 0.006 & 0.079 & 0.112 & 0.268 \\
\midrule
\multicolumn{9}{c}{Three-horizon average} \\
DLinear & 0.2866 & 0.0376 & 0.2399 & 0.0254 & 0.0029 & 0.0823 & 0.0947 & 0.4374 \\
PatchTST & 0.2574 & 0.0385 & 0.2188 & 0.0247 & 0.0029 & 0.0809 & 0.1122 & 0.2792 \\
iTransformer & 0.2695 & 0.0388 & 0.2346 & 0.0267 & 0.0029 & 0.0784 & 0.0991 & 0.2463 \\
\bottomrule
\end{tabular}
\end{table}

\subsection{Bayesian regularization diagnostics}
\label{app:bayesian_diagnostics}
\paragraph{Local adaptation cost.}
We compare Bayesian and prior-free updates at $H=96$, seed 42, on six datasets and five TSFMs. High-change and level-anomaly points are the channel-wise top 10\% of absolute ground-truth first differences and median-absolute-deviation-normalized departures from the median, respectively. Table~\ref{tab:bayesian_local} gives relative MSE changes from plain to Bayesian updates, computed after averaging MSE over the corresponding configurations; positive values indicate a cost.

\begin{table}[htbp]
\centering
\caption{Bayesian regularization cost on selected difficult points ($H=96$, seed 42). MSE changes are in percent.}
\label{tab:bayesian_local}
\small
\setlength{\tabcolsep}{3pt}
\begin{tabular}{@{}lrrrrrrr@{}}
\toprule
Metric & ETTh1 & ETTh2 & ETTm1 & ETTm2 & Exchange & Weather & Overall \\
\midrule
High-change MSE $\Delta$ & +0.20 & -0.03 & -0.51 & +0.28 & +0.89 & +0.03 & -0.09 \\
Level-anomaly MSE $\Delta$ & +0.25 & +0.33 & -0.29 & +1.36 & +0.30 & +0.06 & +0.07 \\
\bottomrule
\end{tabular}
\end{table}

Average local effects are small but vary across datasets.

\paragraph{Directed-prior control.}
Following RegPull in AliO~\citep{park2025alio}, we apply the prior penalty only where the frozen prior predicts the observed target more accurately than the current adapter. Table~\ref{tab:regpull} compares six datasets, five TSFMs, and $H\in\{30,96,336\}$ (90 configurations), with other settings unchanged. MSE is averaged over horizons before computing relative reductions.

\begin{table}[htbp]
\centering
\caption{Relative MSE reduction (\%) with the Bayesian and RegPull-directed priors over three horizons and six datasets. Larger values are better; differences are in percentage points (pp).}
\label{tab:regpull}
\small
\setlength{\tabcolsep}{3pt}
\begin{tabular}{@{}lrrrrrrr@{}}
\toprule
Method & ETTh1 & ETTh2 & ETTm1 & ETTm2 & Exchange & Weather & Overall \\
\midrule
ORCA (Bayesian prior) & 2.68 & 0.26 & 10.04 & 10.56 & 4.00 & 11.52 & 6.51 \\
RegPull-prior & 3.01 & 1.04 & 9.03 & 8.56 & 3.00 & 11.11 & 5.96 \\
Difference (pp) & +0.33 & +0.78 & -1.01 & -2.00 & -1.00 & -0.41 & -0.55 \\
\bottomrule
\end{tabular}
\end{table}

The directed prior improves ETTh1 and ETTh2 but lowers the overall gain from 6.51\% to 5.96\%, supporting the original Bayesian prior. Aggregate MSE alone does not exclude all local over-smoothing effects.

\clearpage
\input{checklist.tex}

\end{document}

%% file: checklist.tex
\section*{NeurIPS Paper Checklist}

\begin{enumerate}

\item {\bf Claims}
    \item[] Question: Do the main claims made in the abstract and introduction accurately reflect the paper's contributions and scope?
    \item[] Answer: \answerYes{}
    \item[] Justification: The abstract and introduction accurately summarize ORCA as a black-box online adaptation framework, and the stated contributions match the method, theory, and experiments reported in the main paper and appendix.

\item {\bf Limitations}
    \item[] Question: Does the paper discuss the limitations of the work performed by the authors?
    \item[] Answer: \answerYes{}
    \item[] Justification: The paper includes an explicit Limitations section, noting that ORCA depends on the base model producing reasonable forecasts and that future work should examine abrupt structural shifts and probabilistic metrics.

\item {\bf Theory assumptions and proofs}
    \item[] Question: For each theoretical result, does the paper provide the full set of assumptions and a complete (and correct) proof?
    \item[] Answer: \answerYes{}
    \item[] Justification: The paper provides full derivations and proofs in the appendix for the context-conditioned learning result, the predictive-space Bayesian update, and the Boltzmann routing regret bound.

    \item {\bf Experimental result reproducibility}
    \item[] Question: Does the paper fully disclose all the information needed to reproduce the main experimental results of the paper to the extent that it affects the main claims and/or conclusions of the paper (regardless of whether the code and data are provided or not)?
    \item[] Answer: \answerYes{}
    \item[] Justification: The appendix discloses the datasets, base models, horizons, split ratio, optimizer, buffer settings, router hyperparameters, and evaluation protocol needed to reproduce the main experiments.

\item {\bf Open access to data and code}
    \item[] Question: Does the paper provide open access to the data and code, with sufficient instructions to faithfully reproduce the main experimental results, as described in supplemental material?
    \item[] Answer: \answerYes{}
    \item[] Justification: The code is available at \url{https://github.com/Fifthky/ORCA}. The appendix describes the datasets, training, evaluation, and hyperparameter settings used for reproduction.

\item {\bf Experimental setting/details}
    \item[] Question: Does the paper specify all the training and test details (e.g., data splits, hyperparameters, how they were chosen, type of optimizer) necessary to understand the results?
    \item[] Answer: \answerYes{}
    \item[] Justification: The paper specifies the chronological 7:1:2 split, look-back window, horizons, optimizer, learning rate, buffer size, batch size, warmup, and router settings, which are the key details needed to understand the results.

\item {\bf Experiment statistical significance}
    \item[] Question: Does the paper report error bars suitably and correctly defined or other appropriate information about the statistical significance of the experiments?
    \item[] Answer: \answerYes{}
    \item[] Justification: Table~\ref{tab:multiseed} reports means and standard deviations across seeds 42, 43, and 44 for three horizons, eight datasets, and five TSFMs (360 runs). Standard deviations quantify seed variability in percentage points; they are not confidence intervals. Appendix~\ref{app:multiseed} defines the aggregation.

\item {\bf Experiments compute resources}
    \item[] Question: For each experiment, does the paper provide sufficient information on the computer resources (type of compute workers, memory, time of execution) needed to reproduce the experiments?
    \item[] Answer: \answerNo{}
    \item[] Justification: The paper provides an efficiency table for a subset of settings, but it does not give complete compute-resource details for all experiments such as total runtime and hardware requirements across the full benchmark suite.
    
\item {\bf Code of ethics}
    \item[] Question: Does the research conducted in the paper conform, in every respect, with the NeurIPS Code of Ethics \url{https://neurips.cc/public/EthicsGuidelines}?
    \item[] Answer: \answerYes{}
    \item[] Justification: This work studies online forecasting adaptation using public benchmark datasets and does not involve human-subject experiments or collection of personal data.

\item {\bf Broader impacts}
    \item[] Question: Does the paper discuss both potential positive societal impacts and negative societal impacts of the work performed?
    \item[] Answer: \answerNo{}
    \item[] Justification: The paper focuses on forecasting methodology and benchmark evaluation rather than a specific deployment, and does not provide a detailed analysis of potential societal benefits and harms in downstream applications.
    
\item {\bf Safeguards}
    \item[] Question: Does the paper describe safeguards that have been put in place for responsible release of data or models that have a high risk for misuse (e.g., pre-trained language models, image generators, or scraped datasets)?
    \item[] Answer: \answerNA{}
    \item[] Justification: The paper does not release high-risk models or data that require special misuse safeguards.

\item {\bf Licenses for existing assets}
    \item[] Question: Are the creators or original owners of assets (e.g., code, data, models), used in the paper, properly credited and are the license and terms of use explicitly mentioned and properly respected?
    \item[] Answer: \answerNo{}
    \item[] Justification: The paper cites the datasets and base models used, but it does not explicitly list license names or terms of use for the existing assets.

\item {\bf New assets}
    \item[] Question: Are new assets introduced in the paper well documented and is the documentation provided alongside the assets?
    \item[] Answer: \answerYes{}
    \item[] Justification: ORCA's method and experimental settings are documented in the paper and appendix. The accompanying implementation is available at \url{https://github.com/Fifthky/ORCA}.

\item {\bf Crowdsourcing and research with human subjects}
    \item[] Question: For crowdsourcing experiments and research with human subjects, does the paper include the full text of instructions given to participants and screenshots, if applicable, as well as details about compensation (if any)? 
    \item[] Answer: \answerNA{}
    \item[] Justification: The paper does not involve crowdsourcing or research with human subjects.

\item {\bf Institutional review board (IRB) approvals or equivalent for research with human subjects}
    \item[] Question: Does the paper describe potential risks incurred by study participants, whether such risks were disclosed to the subjects, and whether Institutional Review Board (IRB) approvals (or an equivalent approval/review based on the requirements of your country or institution) were obtained?
    \item[] Answer: \answerNA{}
    \item[] Justification: The paper does not involve research with human subjects.

\item {\bf Declaration of LLM usage}
    \item[] Question: Does the paper describe the usage of LLMs if it is an important, original, or non-standard component of the core methods in this research? Note that if the LLM is used only for writing, editing, or formatting purposes and does \emph{not} impact the core methodology, scientific rigor, or originality of the research, declaration is not required.
    \item[] Answer: \answerNA{}
    \item[] Justification: The core method does not use LLMs as an important, original, or non-standard component.

\end{enumerate}